\pdfoutput=1

\documentclass[11pt]{article}

\usepackage[]{ACL2023}

\usepackage{times}
\usepackage{latexsym}

\usepackage[T1]{fontenc}

\usepackage[utf8]{inputenc}

\usepackage{microtype}

\usepackage{inconsolata}

\usepackage[normalem]{ulem}
\usepackage{subcaption}
\usepackage{amsmath}
\usepackage[export]{adjustbox}

\usepackage{enumitem}
\usepackage{booktabs}

\newcommand{\draftcomment}[3]{{\textcolor{#3}{[#1]#2}}}
\renewcommand{\draftcomment}[3]{}  

\definecolor{bottlegreen}{rgb}{0.0,0.42,0.31}
\definecolor{donorred}{RGB}{228.,116.,95.}
\definecolor{reciepientblue}{RGB}{0,152,251}
\newcommand{\tomasz}[1]{\draftcomment{#1}{\textsc{tomasz}}{bottlegreen}}

\newcommand{\gabi}[1]{\draftcomment{#1}{\textsc{gabi}}{red}}
\newcommand{\gabis}[1]{\gabi{#1}}

\title{Exploring the Impact of Training Data Distribution and Subword Tokenization on Gender Bias in Machine Translation}

\author{
  Bar Eluz$^{\diamondsuit}$\footnotemark[1] \qquad
  Tomasz Limisiewicz$^{\spadesuit}$\thanks{$\;\;$Equal contribution.}   \thanks{$\;\;$Work partially done while visiting the Hebrew University.} \qquad
  Gabriel Stanovsky$^{\diamondsuit}$ 
  David Mare\v{c}ek$^{\spadesuit}$ \\
  $^{\diamondsuit}\;$School of Computer Science, The Hebrew University of Jerusalem \\
  $^{\spadesuit}\;$Faculty of Mathematics and Physics, Charles University in Prague
 \\
  \texttt{\{bar.iluz,gabriel.stanovsky\}@mail.huji.ac.il} \\
  \texttt{\{limisiewicz,marecek\}@ufal.mff.cuni.cz}}

\begin{document}
\maketitle
\begin{abstract}
We study the effect of tokenization on gender bias in machine translation, an aspect that has been largely overlooked in previous works. 
Specifically, we focus on the interactions between the frequency of gendered profession names in training data, their representation in the subword tokenizer's vocabulary, and gender bias. 
We observe that female and non-stereotypical gender inflections of profession names (e.g., Spanish ``doctora'' for ``female doctor'') tend to be split into multiple subword tokens.
Our results indicate that the imbalance of gender forms in the model's training corpus is a major factor contributing to gender bias and has a greater impact than subword splitting. We show that analyzing subword splits provides good estimates of gender-form imbalance in the training data and can be used even when the corpus is not publicly available.
We also demonstrate that fine-tuning just the token embedding layer can decrease the gap in gender prediction accuracy between female and male forms without impairing the translation quality. \footnote{The code is available at \url{https://github.com/tomlimi/MT-Tokenizer-Bias}} 
\gabis{Does this fine-tuning follow from our previous experiments? If so, how?} \tomasz{It shows the importance of token/lexical representation on Gender bias, }

\gabis{A general thought -- this outline represents how the project evolved chornologically, but a different story may be that we want to test how the training frequency of a profession word affects its gender bias, and tokenization is obviously a major part in that. If we choose this framing then it's less of a ``negative'' result and more of a detailed exploration of the effect of frequency on gender bias? Do we know if other works have looked on this?}
\tomasz{ I agree with the idea. There were a few works observing this association (data -> gender bias), and IMO it's rather intuitive cause. Still, there wasn't a work yet taking into account the interactions of data, tokenizer, and GB. I think that our take-home message is that it's easier (feasible) to post-hoc analyze vocabulary than the training data.}

\end{abstract}

\section{Introduction}
\label{sec:intro}
\begin{figure}[!t]
    \centering
    \includegraphics[width=1.0\columnwidth,center]{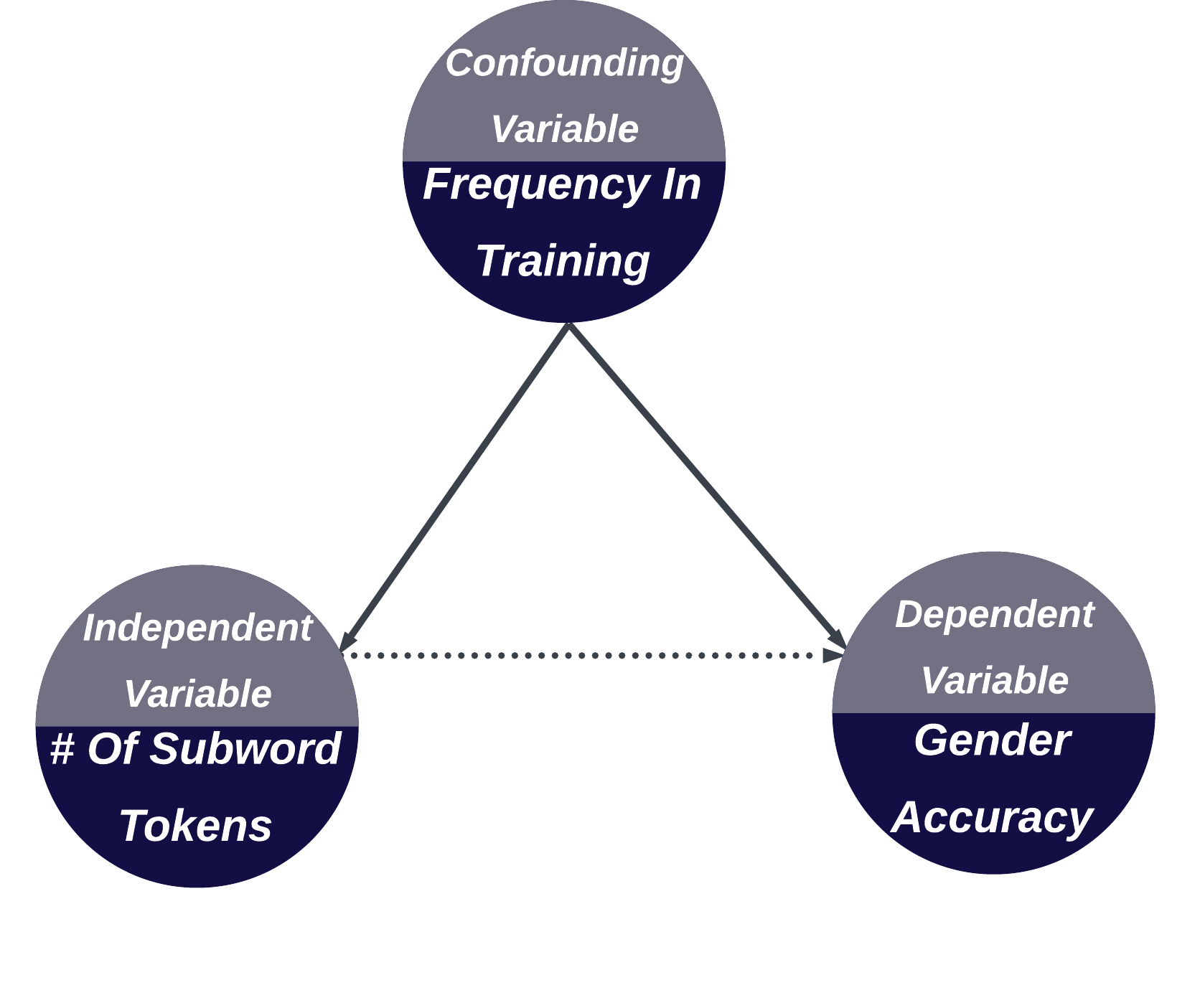}
    \caption{
    The schema depicts two factors affecting the accuracy of translating profession words into correct gender-inflected forms in morphologically rich languages.
    The first factor is the frequency of gender inflections of profession names in the training corpus, and the second is the number of subword tokens that these forms are split into. Our analysis reveals that the frequency significantly correlates with the translation accuracy and number of tokens per word. However, when we control for frequency, the correlation between the number of tokens and translation accuracy is insignificant, indicating that frequency is a confounding variable.
    }
    \label{fig:shwartz}
\end{figure}

Machine translation has been one of the fastest-growing research directions in NLP.
However, with the intensive growth of the technology, multiple potential harms were identified~\cite{hovy-spruit-2016-social}, including gender bias, where models rely on spurious correlations (doctors tend to be male) to make their predictions rather than more meaningful signals in their input~\cite{stanovsky-etal-2019-evaluating}.

There are many reasons for the bias, such as imbalances in the training set or architecture choice. Previous works proposed various approaches to combat gender bias in translation models \cite{saunders-byrne-2020-reducing, escude-font-costa-jussa-2019-equalizing}.


In this paper, we focus on the role of tokenization on gender bias which has been largely overlooked in previous approaches to the problem. 


We want to study the causal relationship between tokenization and gender bias. Specifically, we want to know: 1. How do subword tokenizers handle different gender forms, i.e., are female and non-stereotypical gender forms split into more tokens than male and stereotypical gender forms? 2. Whether subword splitting has an impact on the accuracy of translation? 3. Would subword tokenization's effect be significant when accounting for the frequency of gender forms in the training corpus?

To answer those questions, we analyze pre-trained machine translation models from English to a diverse set of three languages that denote morphological gender in nouns ( German, Spanish, and Hebrew).



First, we compare the number of tokens of different gender forms in the target language and find out that, indeed, female and anti-stereotypical forms are split into more tokens. 
Second, the causality analysis shows that the number of subword tokens may initially appear to explain the translation accuracy of gender forms. However, we find out that these factors are conditionally independent when we also consider the word frequency in the training set, as depicted in Figure~\ref{fig:shwartz}.
To support this finding, we fine-tune the model on gender balanced dataset and update its tokenizer, showing that the dataset's role is more impactful on gender bias than tokenization.



To the best of our knowledge, this work is the first in-depth analysis of interactions between training data, tokenization method, and gender bias. Our findings confirm the previous observations \cite{saunders-byrne-2020-reducing, zmigrod-etal-2019-counterfactual} indicating that the distribution of gender forms in the training data significantly influences the bias of a model. Subword tokenizers, typically trained on the same data, can also perpetuate biases present in the data. We show that it is feasible to analyze the representation of gender forms in the learned vocabulary to obtain information on gender distribution in the model's training corpus even without having access to it. 

\gabis{This seems a bit timid for final paragraph. Anything else we can say beside we support previous observations? Maybe don't lead with that and instead say what's novel?}\tomasz{IMO, the main novelty is the analysis method. Changed the first sentence to underline it!}



\section{Experimental Setup}
\label{sec:resources}


\subsection{Models, languages and tokenizers}
The translation models we used for the analysis are OpusMT~\cite{TiedemannThottingal:EAMT2020},  based on Marian NMT framework \cite{mariannmt}, trained on the Opus dataset\footnote{https://opus.nlpl.eu/}, and \textsc{mBART50}~\cite{tang2020multilingual}, a multilingual encoder-decoder model trained on 50 languages.
Both OpusMT and \textsc{mBART50} use SentencePiece based tokenizer~\cite{kudo-richardson-2018-sentencepiece} with unigram language model~\cite{kudo-2018-subword}. 
We test models in translating from English to German, Spanish, and Hebrew.


\paragraph{Choice of target languages.}
We chose German, Spanish, and Hebrew as target languages since they are diverse, and all assign grammatical gender to profession names, adjectives, and nouns.
Moreover, the authors of this study possess a proficiency ranging from intermediate to native levels in the aforementioned languages.
To highlight typological differences between the languages: Hebrew is a Semitic language from the Afroasiatic language family with Abjad script. 
German is a Germanic language with the Latin alphabet. 
Spanish is a Romance language with Latin script that uses a change of suffix (instead of addition) for male-to-female infections.
Both German and Spanish belong to the Indo-European language family.

\paragraph{Choice of models.}
We chose OPUS and mBART models because they are accessible through Huggingface, they support all the languages we selected for analysis, and they manifest strong performance in translation tasks.
Both models follow state-of-the-art design choices, specifically Transformer architecture and SentencePiece tokenizer, which was shown to be preferred over BPE in multilingual models \citep{bostrom2020byte}.

\subsection{Data}
For the evaluation of gender bias, we use the WinoMT~\citep{stanovsky-etal-2019-evaluating}. It is a synthetic English dataset of sentences containing profession names coreferred by gendered pronouns. The sentences are balanced in terms of the number of male and female pronouns used. The construction of this dataset allows checking if translation models show preference toward particular gender forms. The methods for measuring these preferences are described in the following subsection~\ref{sec:metrics}.

\paragraph{Gender forms in target languages.}
\label{sec:human-translation}

To validate the gender translation correctness of professions, we collected human translations from native speakers of Hebrew, German, and Spanish (three annotators for each language) for a list of 40 professions from WinoBias dataset \cite{zhao-etal-2018-gender}. The list contains an equal share of professions that are majorly performed by men and women (based on labor statistics). Each profession is translated to a pair of masculine and feminine forms with the same stem (e.g., ``Mediziner'' -- ``Medizinerin'' and ``Arzt'' -- ``Ärztin''  are two pairs in German).\footnote{Both pairs are German translations of the English profession ``Doctor''.}
The annotators could propose up to 3 pairs of translations for each profession. Subsequently, the authors selected a list of pairs that were proposed by at least two annotators
\footnote{For detailed rules for combining the final lists and discussion of special cases, see Appendix~\ref{sec:rules_for_combining}}.
As a result, we accepted 67 pairs for German (77\% of the annotators' propositions), 54 pairs for Hebrew (76\% of the propositions),  and 45 pairs for Spanish (73\% of the propositions).
The lists with the translations will be released upon the publication of this work.

\paragraph{Gender forms frequency analysis.}
To estimate the frequency of gender forms in the training corpus, we analyze the OPUS-100 dataset~\citep{zhang-etal-2020-improving}. It is a sample from OPUS collection on which OpusMT was trained. It comprises multiple corpora subjects like movie subtitles, code documentation, and the Bible. We used all three training, development, and test corpora of OPUS-100, which contains 1,004,000 sentences for each of the analyzed languages.

\paragraph{Gender-balanced dataset for fine-tuning.}
\label{sec:ft-dataset}

We compile a  gender-balanced bilingual dataset containing a simple English template: 
``He/She is the [profession]'' paired with translations to Hebrew and German. The number of examples with male and female pronouns is equal. The templates are filled with profession names from WinoMT from English and their translations proposed by the annotators. 

\subsection{Metrics}
\label{sec:metrics}


\paragraph{Gender Translation Accuracy (F1).}
To evaluate the model's performance, we check whether the model translates professions in English from WinoMT dataset to correct gender forms in the target language proposed by annotators.
For instance, the profession ``Physician'', when the pronouns indicate male gender in the source sentence, should be translated to corresponding male forms in German (``Arzt'' or ``Mediziner'').
Respectively, when a pronoun indicates female gender, we should obtain female forms in the translator's output sentence (``Ärztin'' or ``Medizinerin'' for German).
We compute the number of correct occurrences, i.e., translated sentences for which profession and gender match the English source.

We define \emph{recall} for each profession in a specific gender as the share of correct occurrences out of a total number of source (English) sentences where the profession appeared. 

The \emph{precision} for each attested (i.e., proposed by annotators) profession translation is the share of correct occurrences out of the total number of outputted sentences where the profession translation appeared.

In our experiments, we report \emph{F1}, which is the harmonic mean of precision and recall for each attested profession translation.

\paragraph{Measures of Gender Bias (\boldmath$\Delta G$, $\Delta S$, $\Delta T$).}

We use metrics proposed by \citet{stanovsky-etal-2019-evaluating} to measure gender bias in MT: $\Delta G$ measures the difference in gender translation correctness (F1) between masculine and feminine entities; similarly, $\Delta S$ measures the difference in F1 between pro-stereotypical and anti-stereotypical instances of gender role assignments.
\footnote{We use previous works' attribution of stereotype to particular English professions~\citep{zhao-etal-2018-gender}}

We compute $\Delta G$ for each pair of male and female translations.
It is a more fine-grained approach than in previous works:
\begin{equation}
        \Delta G = \text{F1}_{m. trans.} - \text{F1}_{f. trans.}
\end{equation}

Analogically we compute $\Delta S$ as the difference in F1 between pro- and anti-stereotypical translation.
\begin{equation}
        \Delta S = \text{F1}_{pro. trans.} - \text{F1}_{anti. trans.}
\end{equation}

Additionally, we define new metrics that measure the differences in the number of tokens that distinct gender forms split into for each pair of profession name translations.
$\Delta T_G$, analogically to $\Delta G$, quantifies the difference in the number of tokens between the male form and the female form:

\begin{equation}
        \Delta T_G = \text{n. Tokens}_{m. trans.} - \text{n. Tokens}_{f. trans.}
\end{equation}

While, $\Delta T_S$ corresponds to $\Delta S$ and measures the difference in number of tokens between the pro- and anti-stereotypical forms:

\begin{equation}
        \Delta T_S = \text{n. Tokens}_{pro. trans.} - \text{n. Tokens}_{anti . trans.}
\end{equation}

With $\Delta T$ we metricize bias already on the tokenization level and inspect its effect on the machine translation performance.
We expect the words split into more tokens to be harder to predict and thus observe correlations between pairs $\Delta T$ and translation bias metrics: $\Delta G$ and $\Delta S$.


\paragraph{Examples:} In case of translation from English to German, recall for English profession ``Physician'' in female form is the number of times ``Physician'' appeared in the source dataset as female and was translated to ``Ärztin'' or ``Medizinerin'', divided by the number of times ``Physician'' appeared as a female in the source dataset. 
Precision for translation ``Ärztin'' is the number of times ``Physician' appeared in the source dataset as female and was translated to ``Ärztin''  divided by the number of times the word ``Ärztin'' appeared in the translator's output.
F1 for translation ``Ärztin'' is a harmonic mean of recall for the female ``Physician'' and precision for ``Ärztin''.

$\Delta G$ for a pair ``Arzt'' -- ``Ärztin'' is difference in F1 for translation ``Arzt'' and translation ``Ärztin''.
 Because ``Physician'' is a stereotypically male profession $\Delta S$ equals to $\Delta G$.

Both $\Delta T_G$ and $\Delta T_S$ are the difference between the number of tokens the words ``Arzt'' and ``Ärztin'' are divided into by the German tokenizer.
\footnote{Please note that for stereotypically female professions, we have $\Delta G = - \Delta S$ and $\Delta T_G = - \Delta T_S$.}

\section{Experiments and Results}
\label{sec:results}
\begin{figure*}[!t]
    \centering
    \begin{subfigure}{0.32\textwidth}
        \centering
        \includegraphics[width=\textwidth]{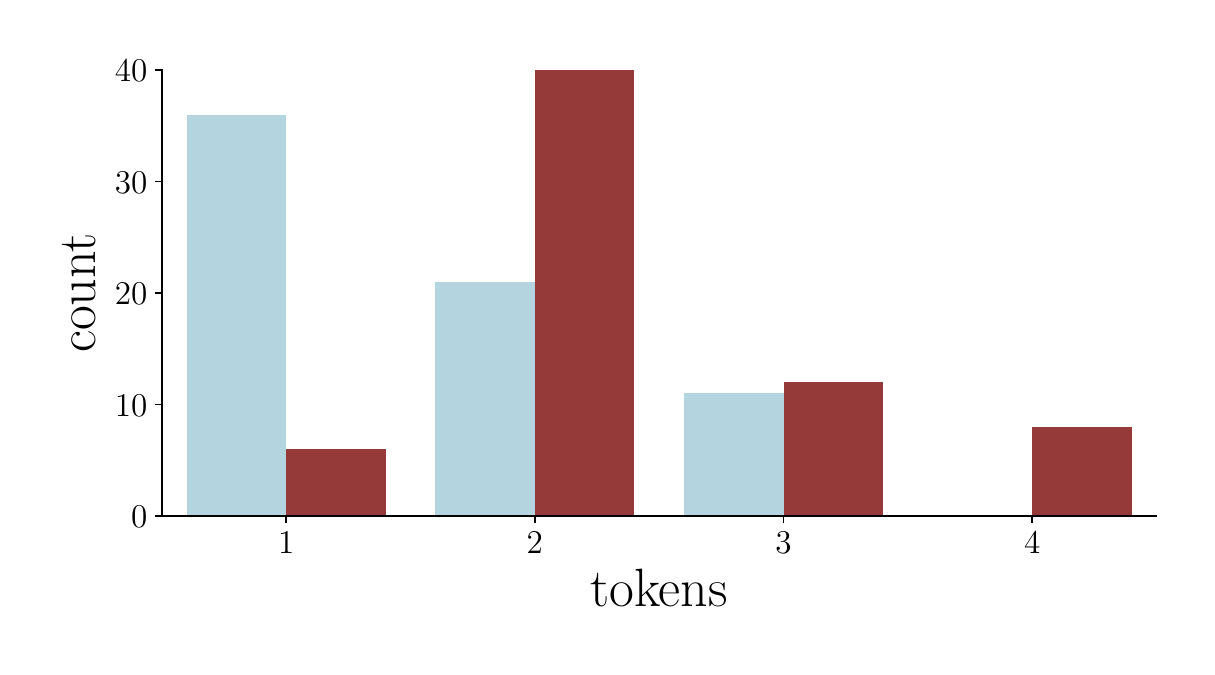}
        \caption{German}
    \end{subfigure}
    \hfill
    \begin{subfigure}{0.32\textwidth}
        \centering
        \includegraphics[width=\textwidth]{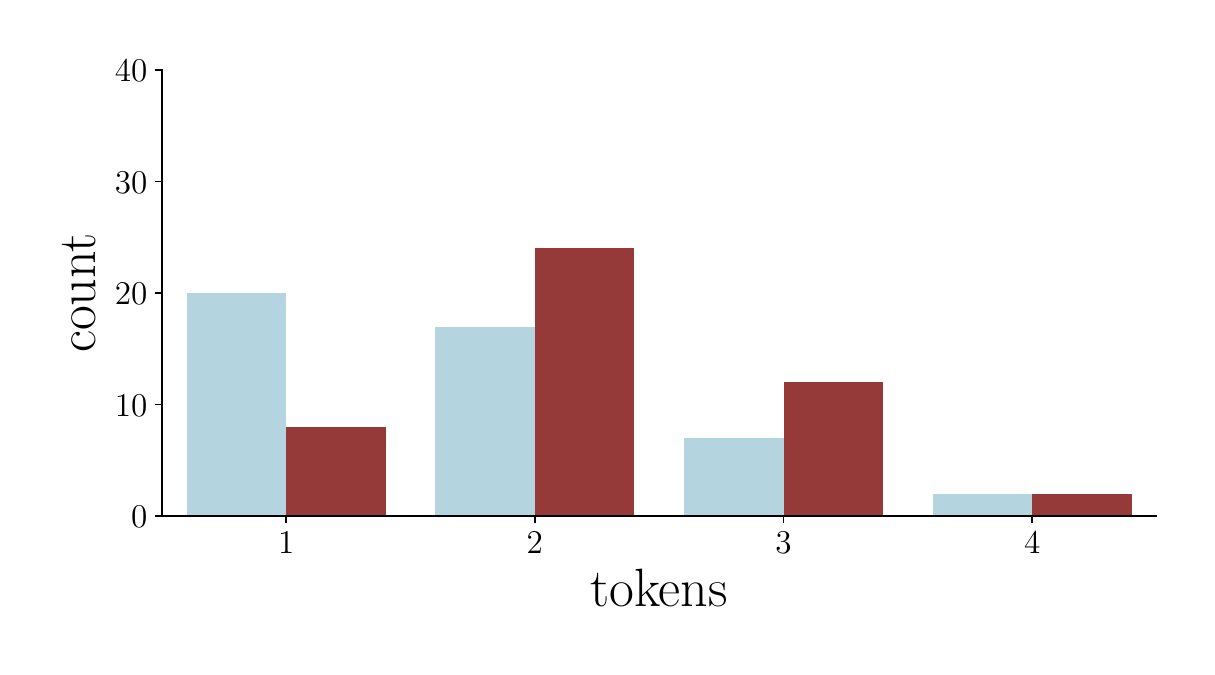}
        \caption{Spanish}
    \end{subfigure}
    \hfill
    \begin{subfigure}{0.32\textwidth}
        \centering
        \includegraphics[width=\textwidth]{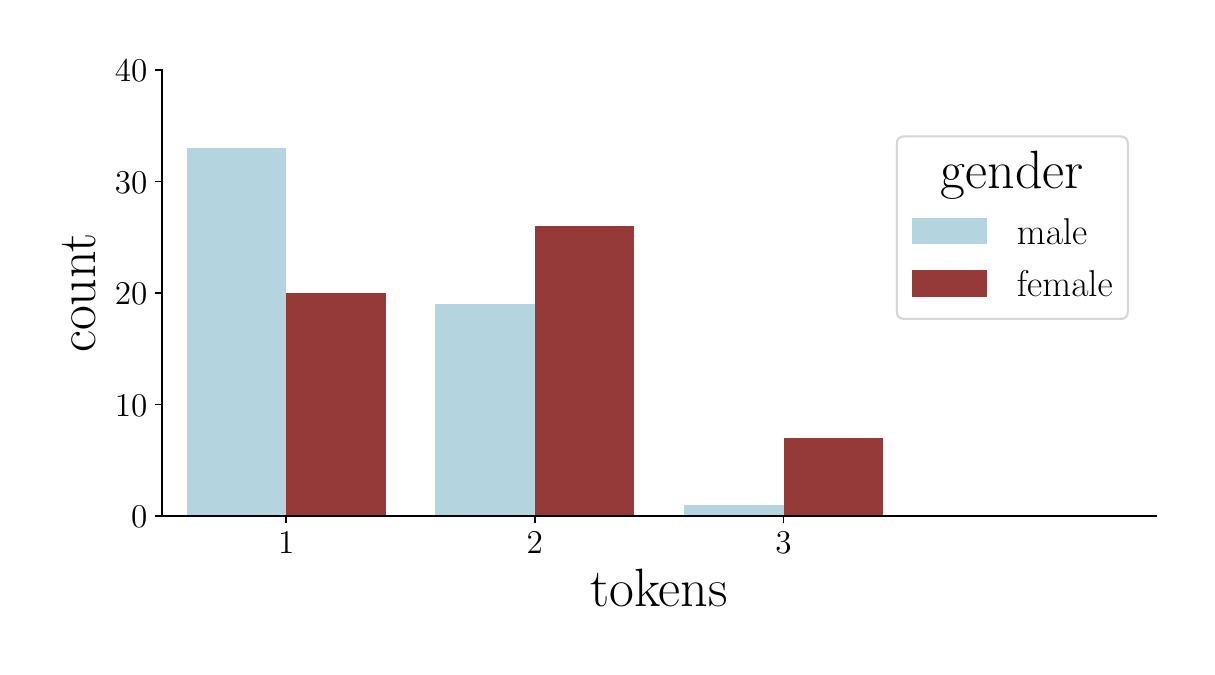}
        \caption{Hebrew}
    \end{subfigure}
    \caption{OpusMT: Human translated profession names were grouped by the number of tokens they were split into. On the x-axis: number of tokens per word. On the y-axis: the count of male and female forms professions in each of the groups. Male forms tend to be split into fewer tokens than female forms.}
    \label{fig:1B_3}
\end{figure*}
\begin{figure*}[!t]
    \centering
    \begin{subfigure}{0.32\textwidth}
        \centering
        \includegraphics[width=\textwidth]{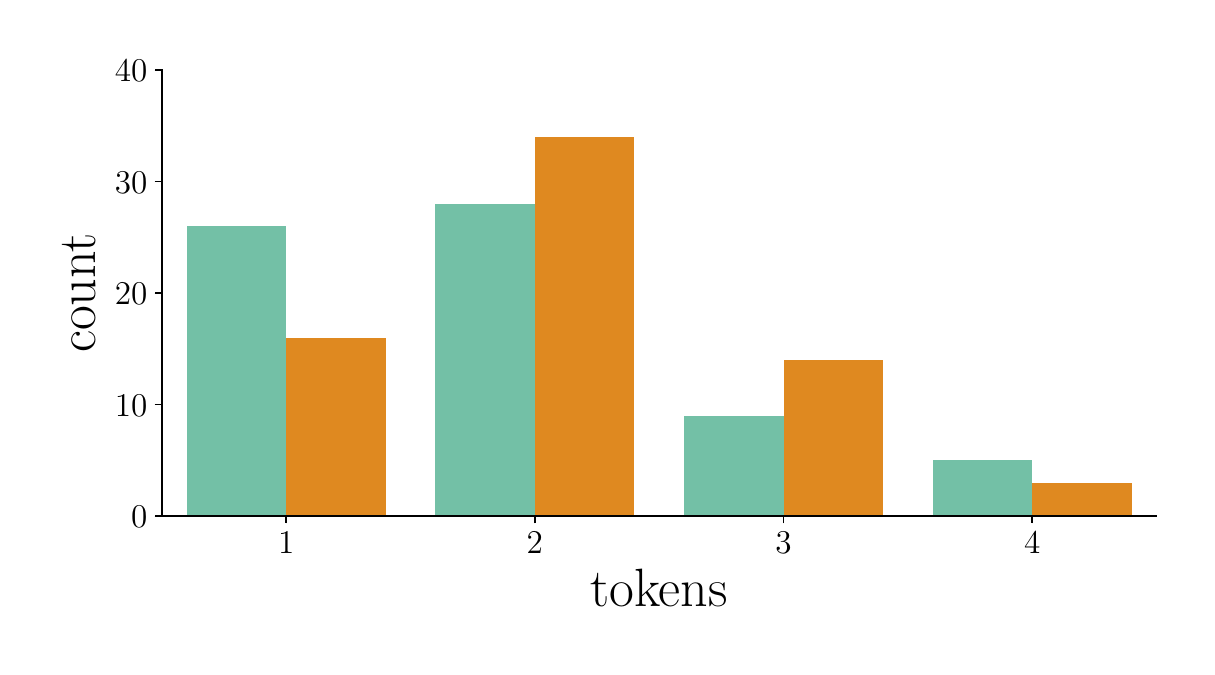}
        \caption{German}
    \end{subfigure}
    \hfill
    \begin{subfigure}{0.32\textwidth}
        \centering
        \includegraphics[width=\textwidth]{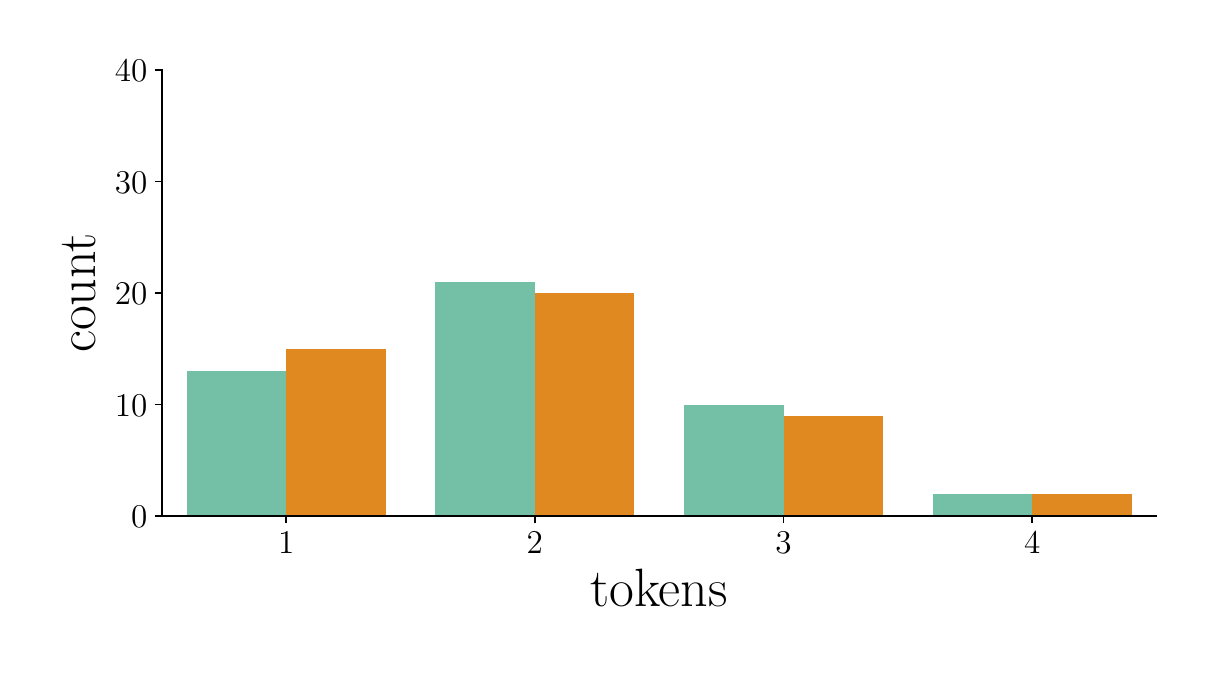}
        \caption{Spanish}
    \end{subfigure}
    \hfill
    \begin{subfigure}{0.32\textwidth}
        \centering
        \includegraphics[width=\textwidth]{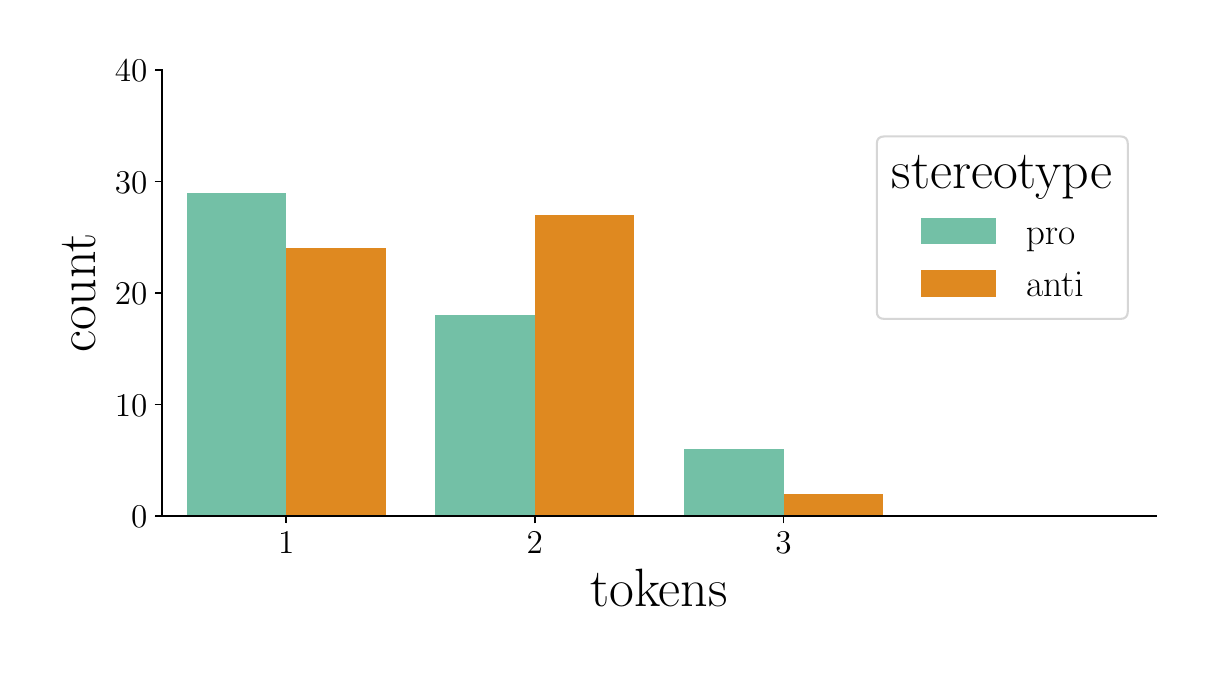}
        \caption{Hebrew}
    \end{subfigure}
    \caption{OpusMT: Human translated profession names were grouped by the number of tokens they were split into. On the x-axis: number of tokens per word. On the y-axis: the count of pro- and anti-stereotypical forms of professions in each of the groups. Pro-stereotypical forms tend to be split into fewer tokens than anti-stereotypical forms.}
    \label{fig:1B_4}
\end{figure*}

To test how tokenization affects the translation accuracy of professionals' gender, and whether a model prefers to generate translations with fewer tokens, we design four experiments explained in the following section.

\subsection{Are female and anti-stereotypical forms split into more tokens?}
\label{sec:1B}

We take human translations (obtained from the procedure described in Section~\ref{sec:human-translation}) and check how many tokens they are split into by the analyzed system's tokenizer.

We expect that female forms will be split into more tokens than male forms, partially due to derivational suffixes appearing only in female forms.

\begin{figure*}[!tb]
    \centering
    \begin{subfigure}{0.32\linewidth}
        \centering
        \includegraphics[width=\textwidth]{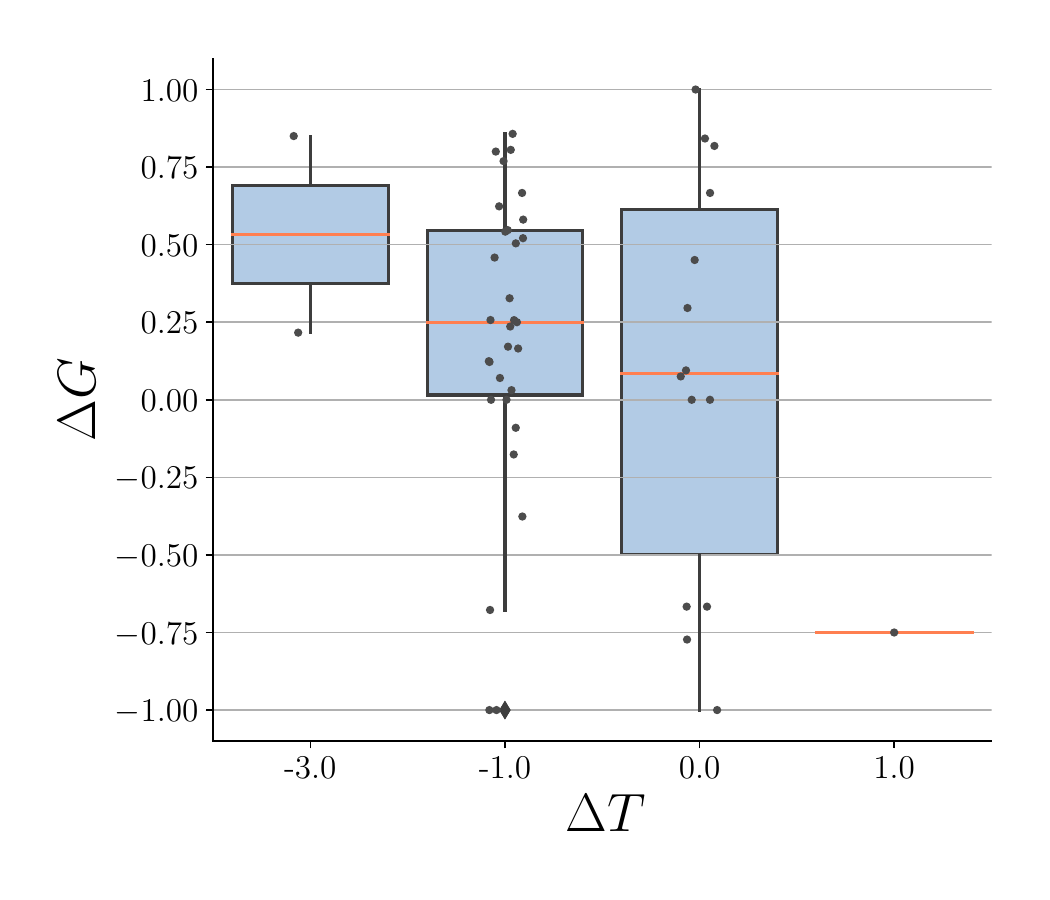}
        \caption{German}
    \end{subfigure}
    \hfill
    \begin{subfigure}{0.32\linewidth}
        \centering
        \includegraphics[width=\textwidth]{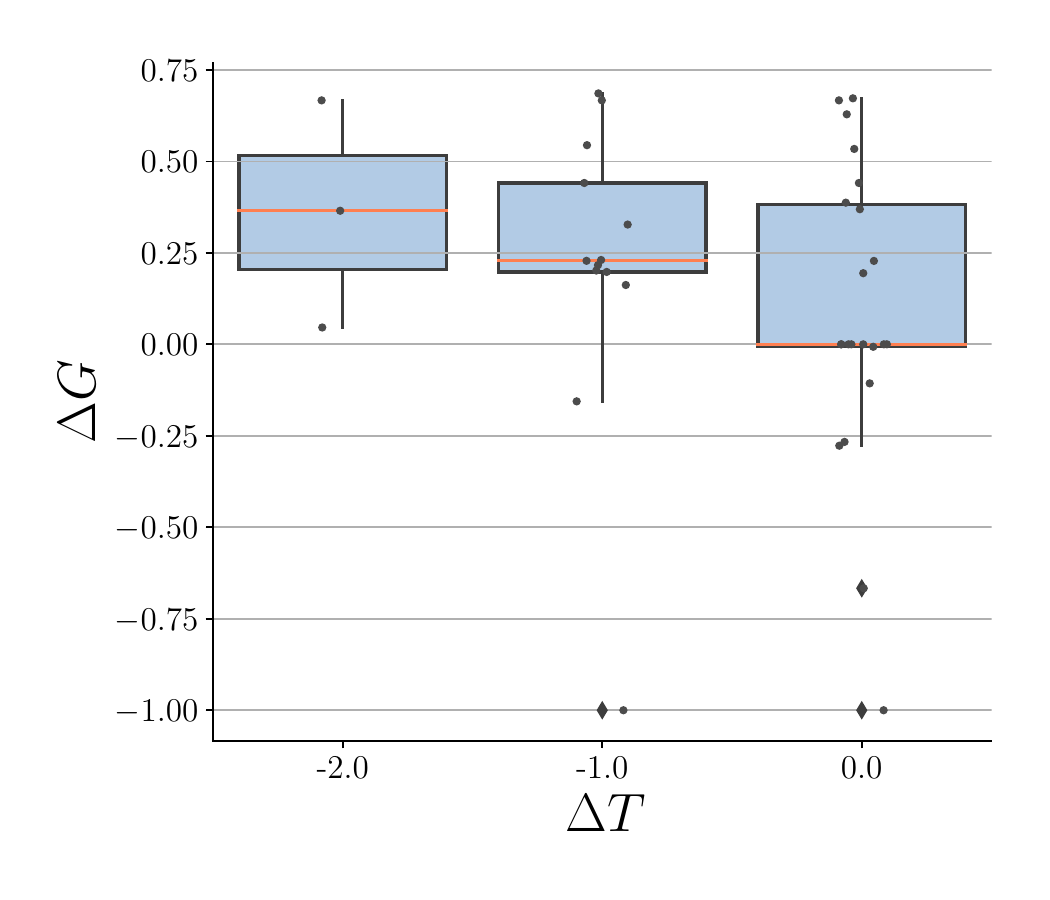}
        \caption{Spanish}
    \end{subfigure}
    \hfill
    \begin{subfigure}{0.32\linewidth}
        \centering
        \includegraphics[width=\textwidth]{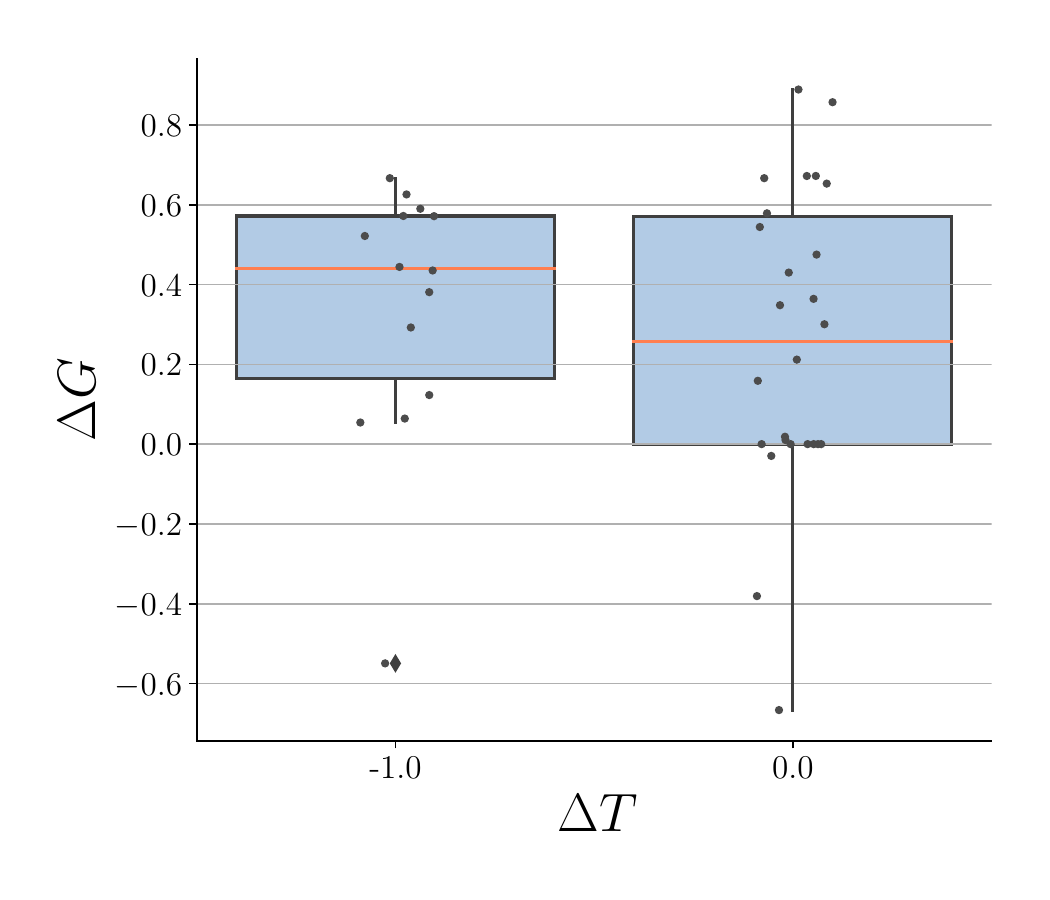}
        \caption{Hebrew}
    \end{subfigure}
    \caption{OpusMT: $\Delta G$ as the difference between F1-score for male and female test instances for each of paired translations. $\Delta T_G$ is the difference between the number of tokens in a male and a female form. An orange line marks the median. Pearson's correlation coefficients and corresponding p-values: $\rho=-0.25$, $p=0.09$ for German, $\rho=-0.21$, $p=0.20$ for Spanish, and  $\rho=-0.11$, $p=0.50$ for Hebrew.}
    \label{fig:1C_delta_g}
\end{figure*}
\begin{figure*}[!tb]
    \centering
    \begin{subfigure}{0.32\linewidth}
        \centering
        \includegraphics[width=\textwidth]{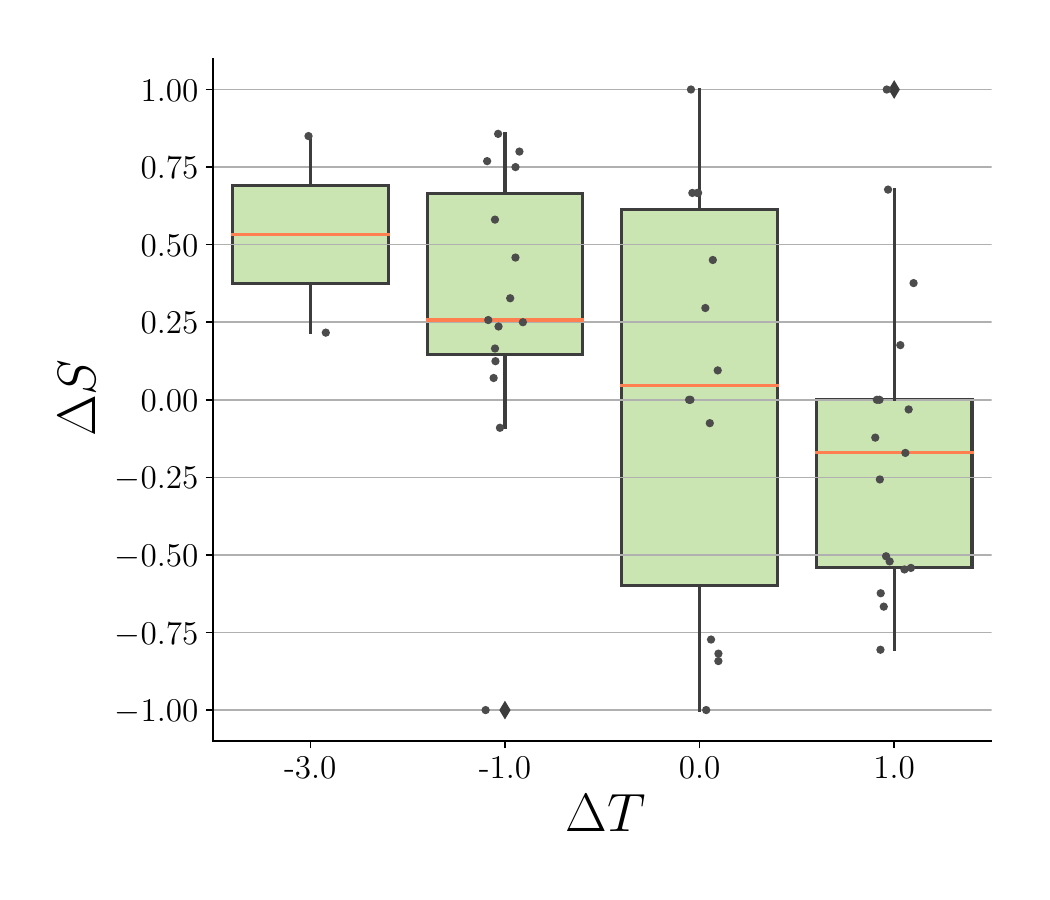}
        \caption{German}
    \end{subfigure}
    \hfill
    \begin{subfigure}{0.32\linewidth}
        \centering
        \includegraphics[width=\textwidth]{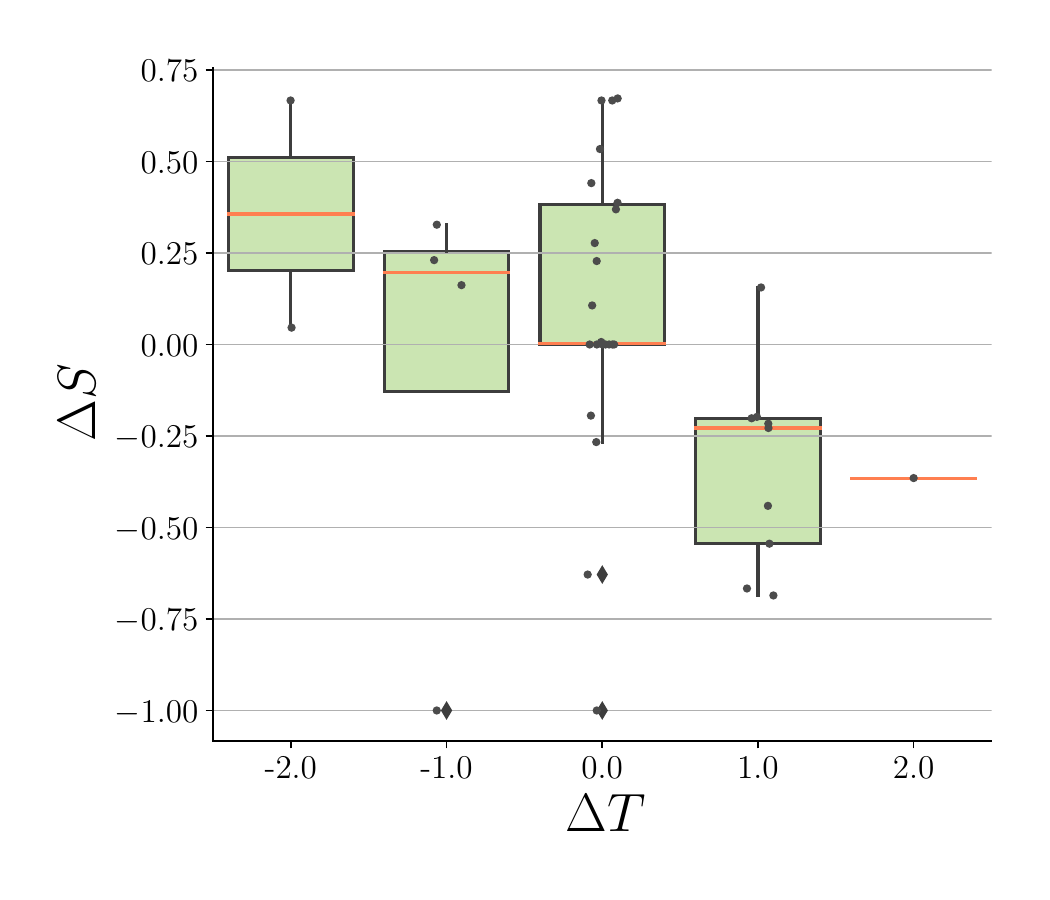}
        \caption{Spanish}
    \end{subfigure}
    \hfill
    \begin{subfigure}{0.32\linewidth}
        \centering
        \includegraphics[width=\textwidth]{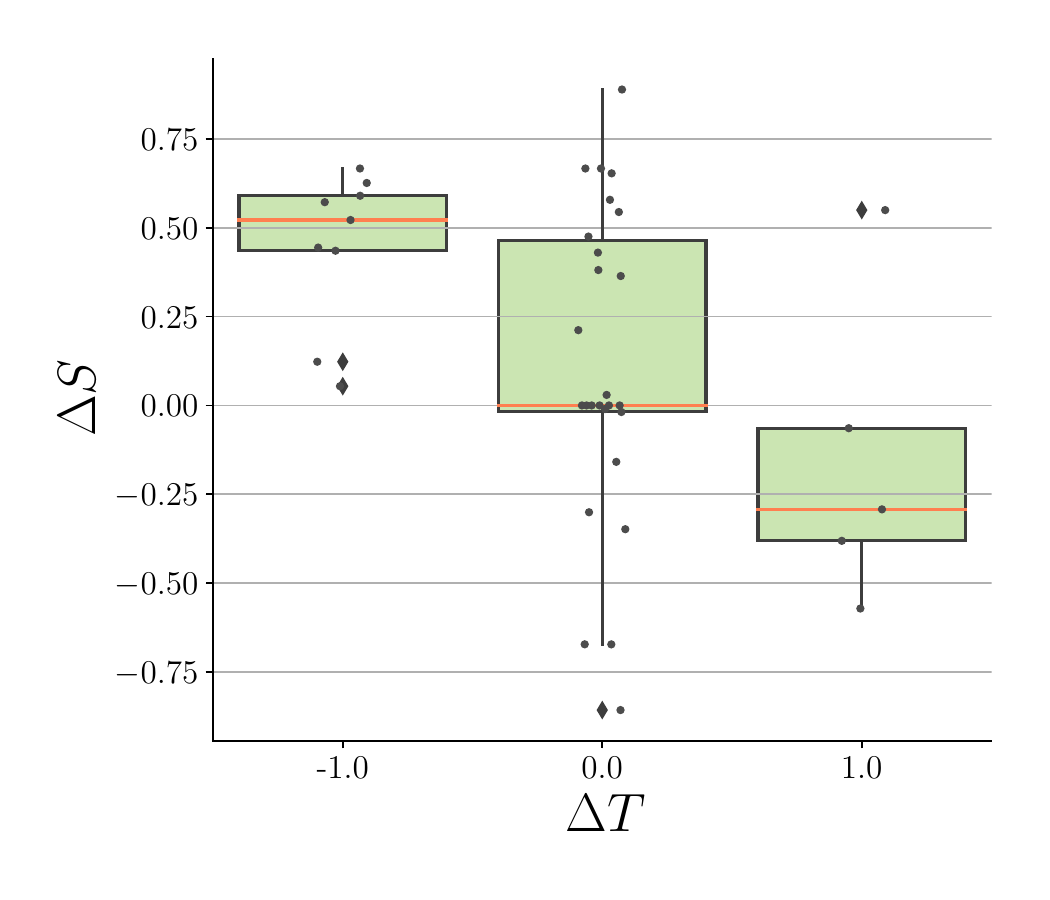}
        \caption{Hebrew}
    \end{subfigure}
    \caption{OpusMT: $\Delta S$ as the difference between F1-score for pro- and anti-stereotypical test instances for each of paired translations. $\Delta T_G$ is the difference between the number of tokens in a pro- and an anti-stereotypical translation. An orange line marks the median. We observe a significant negative correlation between the two measures. Pearson's correlation coefficients and corresponding p-values: $\rho=-0.37$, $p=0.010$ for German, $\rho=-0.37$, $p=0.024$ for Spanish, and $\rho=-0.41$, $p=0.008$ for Hebrew.}
    \label{fig:1C_delta_s}
\end{figure*}

\paragraph{Results}
\label{sec:res-1B}

Figures~\ref{fig:1B_3}~and~\ref{fig:1B_4} show how many translated WinoMT professions are split into a specific number of tokens. We observe that female forms tend to be divided into more tokens than male ones. Only a small portion of female forms are not split.
Similarly, pro-stereotypical translations are split into fewer tokens than anti-stereotypical ones. However, the difference is smaller than in the case of gender.
We observe that the difference in the male and female number of tokens in Spanish is smaller because female forms in Spanish are sometimes expressed by changing the suffix rather than addition (e.g. Consejero vs. Consejera).

The reason why female and anti-stereotypical forms are split into more tokens is probably that they appear in the training corpus less often.\footnote{The results of \textsc{mBART50} are similar to Opus-MT and are represented in Appendix \ref{sec:mbart-results}.}



\subsection{Does subword splitting affect the accuracy of translation?}

We compute bias metrics: $\Delta G$, $\Delta S$ for pairs of translations to compare them with the difference in the number of tokens between gender forms ($\Delta T$). 

We expect that when profession names differ in the number of tokens, the model will more likely generate the shorter form (typically the male or pro-stereotypical). Our inspection is that the preference for shorter forms is connected to gender bias. Therefore we expect to observe a negative correlation between the difference in translation accuracy and the number of subword tokens.

\paragraph{Results}
\label{sec:res-1C}

In Figures~\ref{fig:1C_delta_g}~and~\ref{fig:1C_delta_s}, we observe negative correlations between $\Delta T$ and both $\Delta G$ and $\Delta S$. The relationship is stronger in the latter case. This finding supports our hypothesis that the difference in the number of tokens leads to the model's preference for a form with fewer tokens. Additionally, for translation pairs with $\Delta T=0$, the median of the bias measure distribution is close to zero (with the notable exception of $\Delta G$ for Hebrew). This suggests that the model is less biased for professions when both translation forms are divided into the same number of tokens.

\begin{figure*}[!tb]
    \centering
    \begin{subfigure}{0.49\linewidth}
        \centering
        \includegraphics[width=\textwidth]{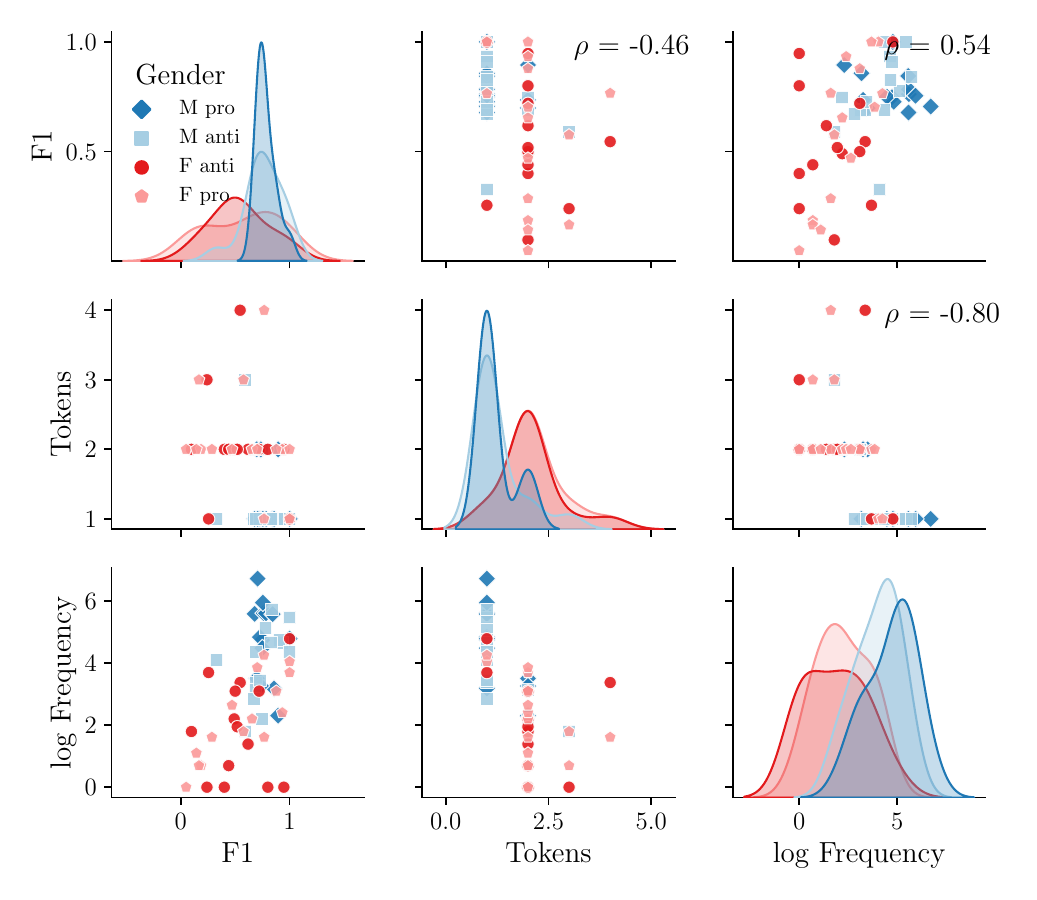}
        \caption{German}
    \end{subfigure}
    \hfill
    \begin{subfigure}{0.49\linewidth}
        \centering
        \includegraphics[width=\textwidth]{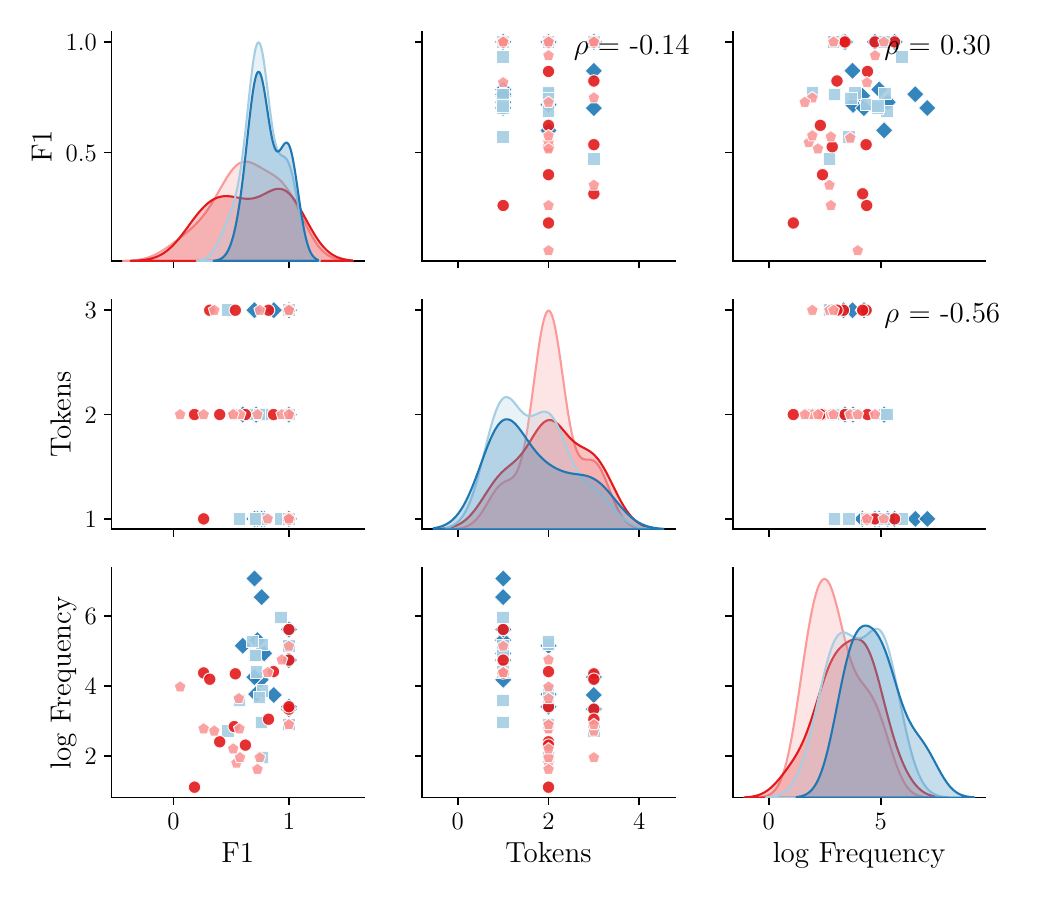}
        \caption{Spanish}
    \end{subfigure}
    \vfill
    \begin{subfigure}{0.49\linewidth}
        \centering
        \includegraphics[width=\textwidth]{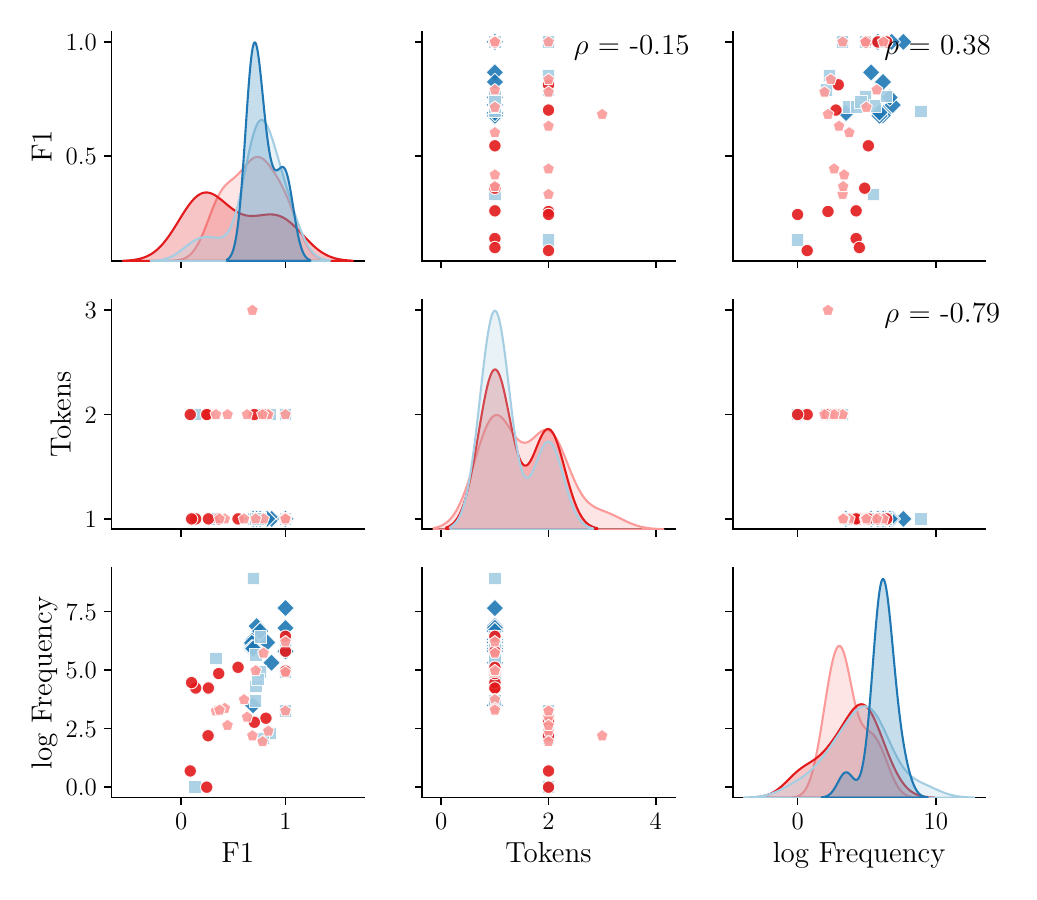}
        \caption{Hebrew}
    \end{subfigure}
    \caption{Pair analysis of gender prediction performance (F1), number of tokens, and the frequency of each profession in the OPUS-100 dataset. Each style of dots represents professions in the male/female pro-stereotypical/anti-stereotypical form. The diagonal plots show the density of the feature for the specific gender and stereotype sets.}
    \label{fig:1D}
\end{figure*}

\subsection{What is the causal relationship between tokenization, training data, and gender prediction accuracy?}

An alternative explanation of the negative trend observed in the previous experiment is the presence of an underlying factor, in our case, the frequency of specific gender forms in the training corpus. Previous research has shown that the terms' frequency affects both tokenization \cite{kudo-2018-subword} and gender bias \cite{escude-font-costa-jussa-2019-equalizing}.


In this experiment, we measure the significance of the correlation between those three factors. Also, we check the conditional independence between the number of tokens per target profession and gender prediction accuracy (measured by F1 score for each target profession described in Section~\ref{sec:resources}) given the profession form's frequency in the training corpus:


\begin{equation}
    F1 \perp \text{n. Tokens} \mid \text{Frequency}
\end{equation}
\label{eqn:conditional-indipendence}

\paragraph{Results}
\label{sec:res-1D}

In Figure~\ref{fig:1D}, we observe that as the number of tokens decreases, F1 increases.
Taking into account the correlation coefficient, the F1 measure is more sensitive to the frequency of a word in the training corpus.
Moreover, less frequent words tend to be split into more tokens. From the density plots (on the diagonal of the figure), we see that male profession words (especially pro-stereotypical ones) appeared much more often in the training corpora. Thus, they tended to be split into fewer tokens.

All the correlations between frequency and the two remaining factors are statistically significant ($p < 0.05$), while the correlation between the number of tokens and F1 score is significant only for German.

We performed Jonckheere-Terpstra test \cite{Jonckheere1954ADK} to check conditional independence as described in Equation~\ref{eqn:conditional-indipendence}. The test showed that conditional independence cannot be rejected for all the target languages ($p=0.78$  for German, $p=0.62$ for Spanish, and $p=0.39$ for Hebrew). Those results show that the frequency of a word is a confounding factor affecting both the number of tokens per profession word and F1 scores. \textbf{Hence subword splitting of profession words is not a significant contributor to the correctness of gender prediction given the frequency of gender forms in the training corpus.}

\begin{table*}[!t]
\centering
\begin{tabular}{@{}l|ccc|ccc|ccc|ccc@{}}
\toprule
         & \multicolumn{3}{|c}{Accuracy} & \multicolumn{3}{|c}{$\Delta$ G} & \multicolumn{3}{|c}{$\Delta$ S} & \multicolumn{3}{|c}{BLEU} \\ \midrule
         & DE       & ES      & HE      & DE       & ES       & HE       & DE       & ES       & HE       & DE     & ES     & HE     \\ \midrule
original & 62.2     & 56.3    & 52.1    & 9.6      & 16.4     & 12.3     & \bf12.5     & \bf16.7     & 40.4     & \bf31.4   & \bf42.8   & \bf36.0   \\
FT       & 67.1     & \bf58.9    & \bf54.0    & -2.9     & \bf10.9     & \bf7.8      & 21.5     & 20.0     & \bf35.1     & 31.3   & 42.0   & 35.8   \\
FT + VU  & \bf67.8     & 58.0    & 52.4    & \bf1.1      & 14.0     & 10.4     & 21.7     & 18.2     & 40.7     & 29.1   & 38.3   & 35.1   \\ \bottomrule
\end{tabular}
    \caption{Embedding fine-tuning results in German and Hebrew. Original OpusMT model results compared with the model after embedding layer fine-tuning (row 2) and the model after embedding layer fine-tuning and updating vocabulary with all profession gender forms in target language (row 3).}
    \label{tab:debias-german-hebrew}
\end{table*}

\subsection{Will intervening in the model's training data and the tokenizer's vocabulary reduce gender bias?}

To verify our findings about causes of gender bias, we propose two interventions in the translation models to German, Spanish, and Hebrew: 1. fine-tuning on a dataset with an equal number of male and female forms;\footnote{The dataset contains the sentences in the form ``He/She is the [profession]'' translated to the target language and filled with professions from WinoMT.} 2. Adapting the tokenizer's vocabulary by adding all translations proposed by annotators to assert that they will not be split into subword tokens.


We monitor standard gender bias measures from \citet{stanovsky-etal-2019-evaluating}: gender accuracy, $\Delta G$, and $\Delta S$,\footnote{In this experiment, we use the original formulation of  $\Delta G$ and $\Delta S$. It means $F1$ is computed for the whole dataset, instead of particular professions as in the previous experiments.} and also BLEU on OPUS-100 test split to check if fine-tuning leads to deterioration of translation quality. To determine if the potential improvement results from embeddings' fine-tuning or adding profession words to the vocabulary, we evaluate the baseline, where the embedding layer is fine-tuned without updating the tokenizer's vocabulary.

\paragraph{Results}

Table~\ref{tab:debias-german-hebrew} shows that fine-tuning embedding layers improve the accuracy of translating to the correct gender and decrease preference of male forms ($\Delta G$), while the quality of translation stays on a similar level as in the original model. Protecting gender forms from splitting (by adding them to the tokenizer's vocabulary) only slightly further converges $\Delta G$ towards zero for German while bringing over 2 points drop in BLEU. Performing the vocabulary update before
fine-tuning models deteriorates bias measures and BLEU for Hebrew. For Spanish, $\Delta S$ improves while $\Delta G$ and BLEU worsen.

Interestingly, fine-tuning the embedding layer increases the stereotypical bias ($\Delta S$) for German and Spanish and decreases it only slightly for Hebrew, suggesting that this method could not be sufficient for mitigating this type of bias.

The results confirm our previous observation that the training (or fine-tuning) dataset has more impact on gender bias encoded in the model than subword splitting.


\section{Related Work}
\label{sec:background}
\paragraph{Evaluation and Mitigation of Gender Bias in NMT}

\citet{escude-font-costa-jussa-2019-equalizing} created a set of sentences of the format "I've known [him/her/Mary/John] for a long time, my friend works as a [profession]."
They translated all sentences to Spanish and checked whether "friend" was translated as "amigo" (male)  or "amiga" (female).
They found out that "Him" is predicted at almost 100\% accuracy for all models, but the accuracy
drops when predicting the word "her" on all models.

\citet{stanovsky-etal-2019-evaluating} created a mechanism to evaluate gender bias in machine translation. They showed that almost all translation systems perform significantly better on male roles.

The past works acknowledged the effect of gender form imbalance in training corpus on gender bias manifested in the models. Specifically, \citet{zmigrod-etal-2019-counterfactual} reduce the bias by training the NMT systems on data augmented by adding female forms. \citet{saunders-byrne-2020-reducing, costa-jussa-de-jorge-2020-fine} propose a debiasing algorithm based on fine-tuning the model on a dataset with a comparable number of male and female forms. These approaches are more sustainable because they do not require training the model from scratch. These approaches are in line with our observation that the imbalance of gender forms in training data is the key source of gender bias in the model.

\citet{savoldi-etal-2021-gender-bias} survey the methods for evaluating and mitigating bias in machine translation, identify risks connected to them and propose direction for their improvement.

\citet{guo-etal-2022-auto} propose an automatic prompt-based method to mitigate the biases in pretrained language models. They identify biased prompts and propose a distribution alignment loss to mitigate it.


\paragraph{The Role of Tokenization on System's Performance}

\citet{Domingo2018HowMD} show that tokenizers play a significant role in the neural machine translation pipeline. The tokenization of the target language affects evaluation measures (BLEU) by up to 20 points.

\citet{bostrom-durrett-2020-byte} compare the popular subword tokenizers: Unigram \cite{kudo-2018-subword} and BPE \cite{sennrich-etal-2016-neural}. They show that the former more often splits words on morphological boundaries and thus can improve the model's performance on downstream tasks.

While the recent survey \cite{mielke-etal-2021-between} shows that the choice of effective tokenization method depends on the task, and there is no specific tokenization algorithm suiting all applications.

The role of tokenization on gender bias was not widely evaluated in past works. \citet{libovicky-etal-2022-dont} showed that character-based translators to morphologically rich languages (Czech and German) obtain similar bias results as subword-based systems, even though female forms in these languages contain relatively more characters. It aligns with our observation that tokenization is not a significant source of bias.

\citet{gaido2021split} explored how segmenting methods influence gender bias in speech translation models. Such models have different features, such as vocal characteristics used to measure gender bias over all words, while our work focuses on gender bias in profession words.
In contrast to their work, we focus on formal gender bias definitions and analyzing the relation between frequency, accuracy, and number of tokens.


\section{Discussion and Future Work}
\label{sec:discussion}


Our confirms the validity of the causal schema depicted in Figure~\ref{fig:shwartz} explaining the lower translation accuracy for female and anti-stereotypical profession names. We consciously analyzed the dataset of only non-ambiguous sentences where the gender of each profession is known. This selection was made to enable us to evaluate the accuracy of the ground-truth gender.

Furthermore, the stereotypical occupations are based on US Department of Labor statistics, but it cannot be guaranteed that these same stereotypes are present in other cultures.  
However, it can be considered a reasonable estimate for other languages, as evidenced by the observation that non-stereotypical professions appear less frequently in their training corpora.
Future research may analyze gender roles in target languages to corroborate these observations.

Another future point of interest is mapping more factors for bias by isolating more features.
The dependencies of tokens, frequency, and gender bias will be examined on a larger scale, with more words and different types of tokenization.
We also intend to broaden the scope of the analysis to other languages and include neutral gender forms for the already analyzed languages.

\section{Conclusions}
\label{sec:conclusions}
Our study found that profession words in German, Hebrew, and Spanish tend to be split into more tokens in the female form than in the male form. We then investigated whether this phenomenon amplifies the tendency of NMT models to translate male professions more accurately than female ones. Our results showed that the frequency of gender forms confounds the relationship between the number of tokens and gender bias in the training set. However, the number of subword tokens per word can be used to estimate its frequency in the unavailable training corpus.

The findings of our analysis of translation models were supported by the trends observed in the results of the proposed debiasing method. Specifically, we found that fine-tuning token embeddings on a gender-balanced dataset had a more significant impact on reducing bias than updating the tokenizer's vocabulary with underrepresented gender forms.


These findings suggest that future research should also focus on other aspects of NMT models in order to mitigate gender bias effectively.






\section*{Acknowledgments}
\label{sec:acknowledge}
We thank our annotators for their immense help in collecting data: Johannes Knaute, Kathy Hämmerl, Leonie Weissweiler, Uri Berger, Daniel Rotem, Almog Guriel, Itamar Katz, Jonathan Posternak.

We also thank anonymous reviewers for their valuable comments on the previous versions of this article.  
This work was partially supported by the Israeli Ministry of Science and Technology (grant no. 2088) and by grant 23-06912S of the Czech Science Foundation.
Tomasz Limisiewicz's visit to the Hebrew University has been supported by grant 338521 of the Charles University Grant Agency and the Mobility Fund of Charles University.
We have been using language resources and tools developed, stored, and distributed by the LINDAT/CLARIAH-CZ project of the Ministry of Education, Youth and Sports of the Czech Republic (project LM2018101).

\section*{Limitations}
\label{sec:limitations}
As pointed out in the last section, tokenization can reinforce gender bias in the models but is not the sole factor responsible for gender bias. We expect that even if male and female forms were tokenized to the same number of tokens, bias could still persist in the models.

The important aspect is the frequency of words in the training corpus, which also affects tokenization. We did not have exact information on the model's training corpus. However, the analysis of the OPUS-100 dataset, which was sampled from the full dataset, showed that male forms are much more frequent than female ones.

In addition, we analyzed only 40 stereotypical professions from WinoMT, which is only a subset of the whole dataset. Extending the analysis to other professions without typical gender-role assigned can reveal slightly different trends.

Finally, non-zero bias measures indicate the presence of bias in the model, but the opposite implication is not true. Even in the case where both $\Delta G$ and $\Delta S$ are close to zero, we cannot claim that the model is unbiased, as there may be other bias manifestations of bias involved.

A significant limitation of this work is that we compare only male and female forms of the professions while neglecting neutral forms. We expect neutral forms to be even harder for the model to predict than female forms because of the added suffixes that are rarely present in the training data. For instance, German uses forms with ``*'' and female prefixes (e.g., ``Mechaniker*in'') while ``\char`\\'' is used in Hebrew for that purpose.
Our initial analysis shows that neutral forms are sporadic in the training corpora. For instance, in German, only 7 neutral forms are using the inclusive suffix ``*in'', compared to 1351 female forms of the analyzed professions. The proportion of neutral:female:male forms is approximately 1:193:1845.

\section*{Ethics Statement}
We introduce new methods for evaluating and mitigating gender  bias in machine translation. We believe that this solution will incentivize the development of fairer models in the future.

We do not identify any ethical risks connected to this work.

\bibliography{custom}
\bibliographystyle{acl_natbib}
\clearpage

\appendix
\section{Human Translation}
\label{sec:rules_for_combining} 

\subsection{Post-processing of Collected Data}

The post-processing of the human translations was done according to the following rules:
First, we kept the translations suggested by at least two translators. If there’s no translation option suggested at least two times, we decided which translations to keep according to our knowledge of the language. This happened only for the following translations in Hebrew: ``attendant'', ``construction worker'', and ``analyst''. The translation selection for those words was made by an author who is a native Hebrew speaker.
For professions where translators propose the same  word as both male and female translations, we keep those as valid pairs.

In case spelling is inconsistent among annotators, we kept the more common spelling while fixing pronunciation mistakes according to our knowledge.
Lastly, we keep at most five pairs of translations per profession.

\subsection{Instructions for Annotators}

The goal of the task is to get the correct translation of profession words into your language for both Male and Female forms. 

\begin{enumerate}
\item Select the language you want to translate to (your native language: German, Spanish, or Hebrew).

\item For each word in English, write up to 3 translations both in Male and Female form. 

\item Organize the translation into pairs that differ only in endings 
(e.g., words: Berater and Beraterin are a pair but Beraterin and Ratgaber are not a pair in German). 
Put these words in the neighboring cells (“Translation N Male" and "Translation N Female").

\item in case there is no translation in the other gender that differ only in an ending, leave the other cell of the pair blank.

\item If you propose less than 3 translations in each gender, leave the last cells in a row empty.

\end{enumerate}


\section{Details of Fine-tuning}
\label{sec:fine_tuning_details} 
In fine-tuning, we train only the translation model's input/output embedding layer while keeping the rest of the parameters frozen. We further trained the model for three epochs on the gender-balanced dataset (Section~\ref{sec:ft-dataset}) with batches consisting of 16 examples. For optimization, we used Adam \cite{kingma-2015-adam} with default parameters with a constant learning rate of $5e-5$. The fine-tuning took about 10 minutes on Nvidia A40 GPU.

\section{\textsc{mBART} results}
\label{sec:mbart-results}

\begin{figure*}[!t]
    \centering
    \begin{subfigure}{0.32\textwidth}
        \centering
        \includegraphics[width=\textwidth]{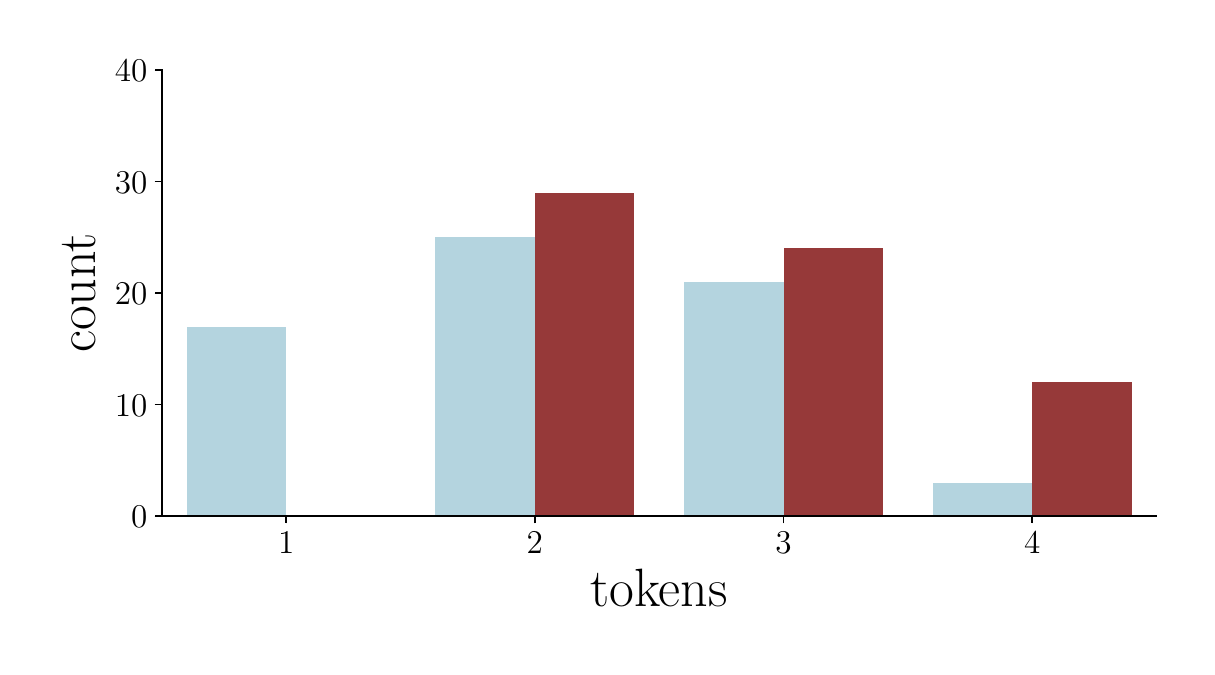}
        \caption{German}
    \end{subfigure}
    \hfill
    \begin{subfigure}{0.32\textwidth}
        \centering
        \includegraphics[width=\textwidth]{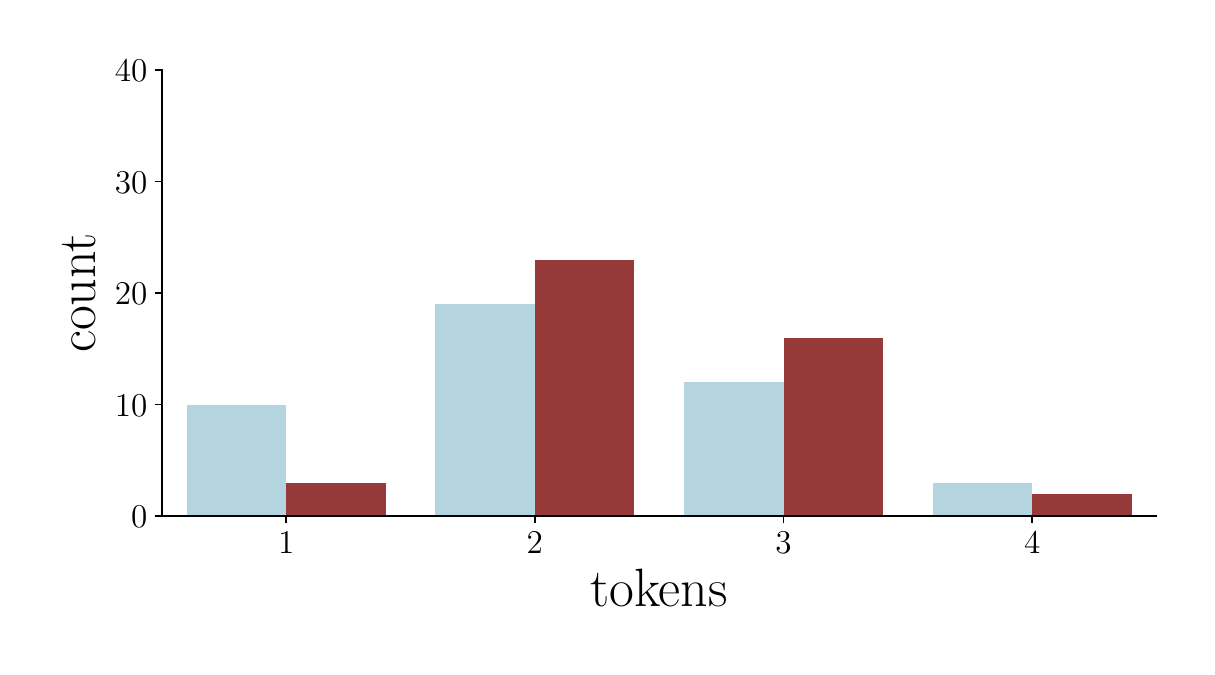}
        \caption{Spanish}
    \end{subfigure}
    \hfill
    \begin{subfigure}{0.32\textwidth}
        \centering
        \includegraphics[width=\textwidth]{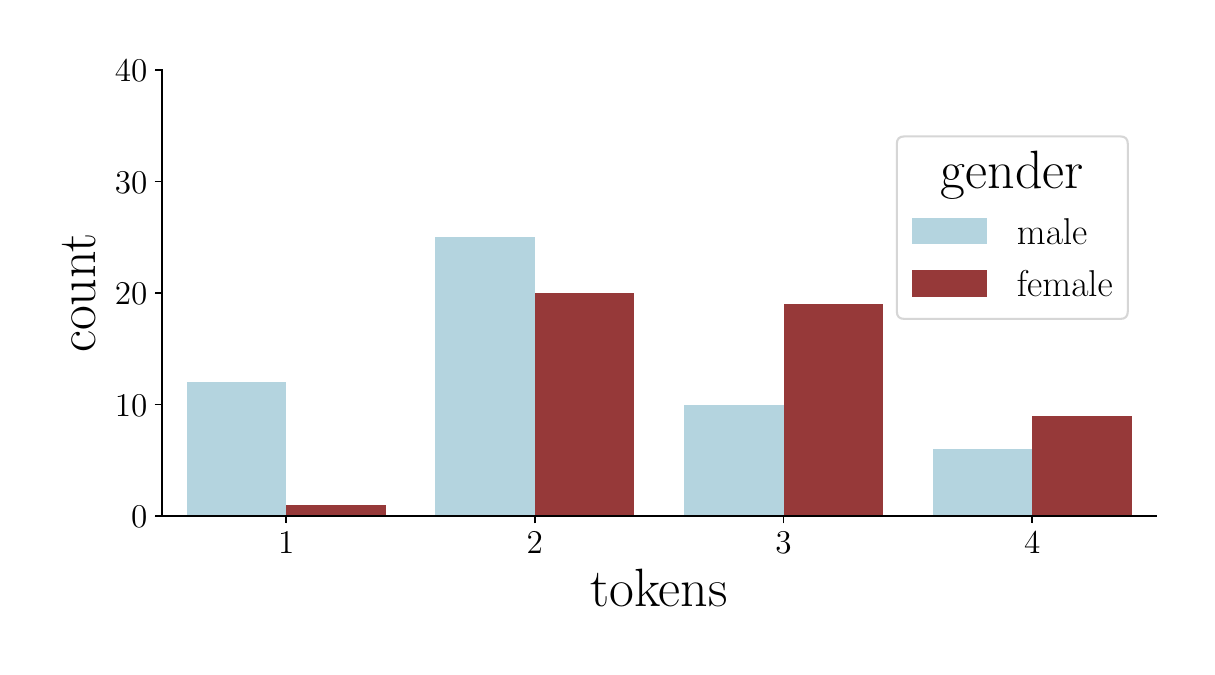}
        \caption{Hebrew}
    \end{subfigure}
    \caption{\textsc{mBART}: Human translated profession names grouped by the number of tokens they were split into. On the x-axis: number of tokens per word. On the y-axis: the count of male and female forms professions in each of the groups. Male forms tend to be split into fewer tokens than female forms.}
    \label{fig:1B_3_mbart}
\end{figure*}
\begin{figure*}[!t]
    \centering
    \begin{subfigure}{0.32\textwidth}
        \centering
        \includegraphics[width=\textwidth]{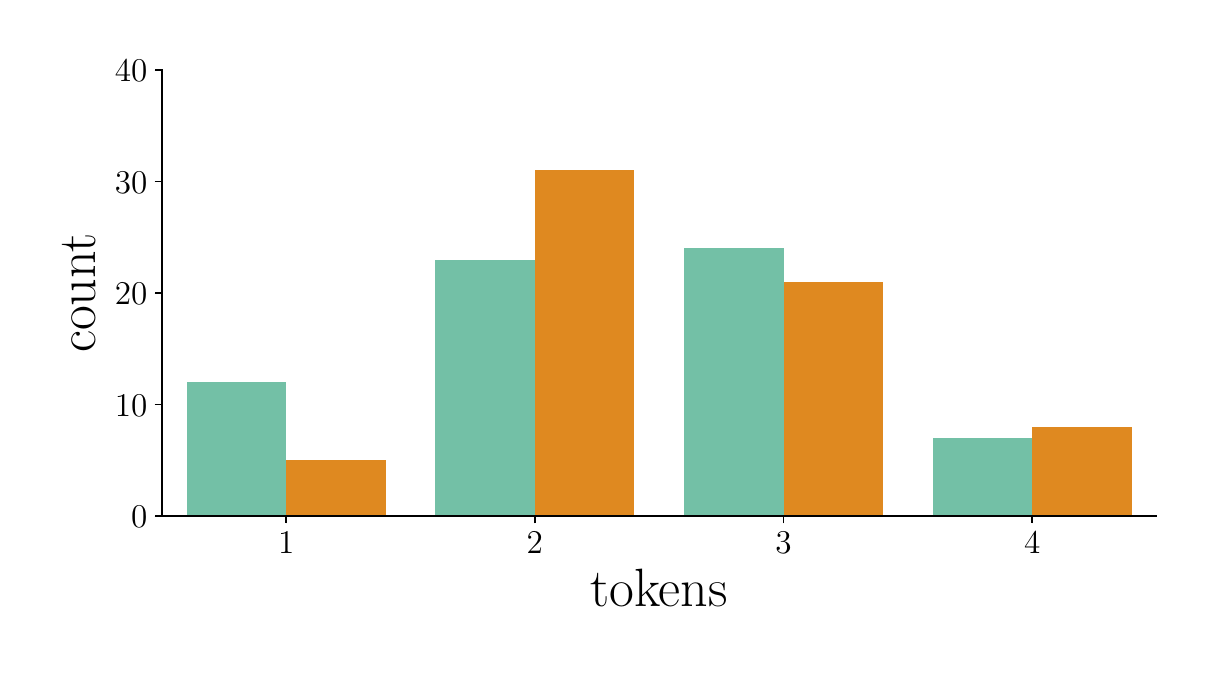}
        \caption{German}
    \end{subfigure}
    \hfill
    \begin{subfigure}{0.32\textwidth}
        \centering
        \includegraphics[width=\textwidth]{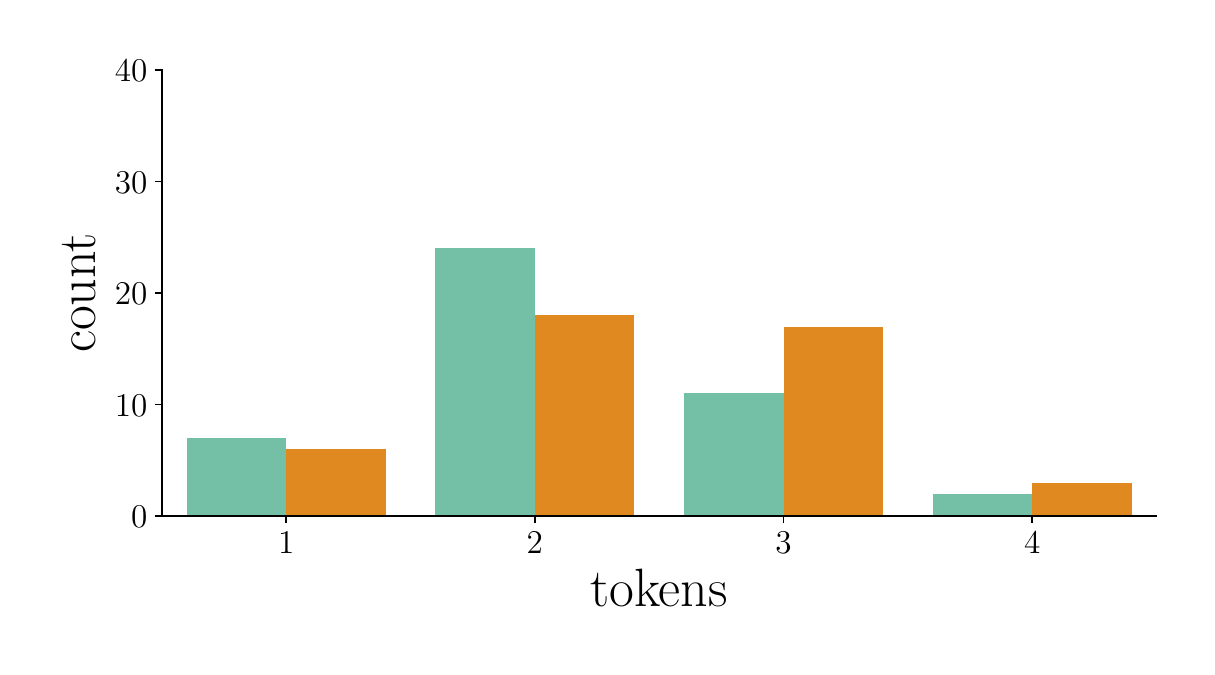}
        \caption{Spanish}
    \end{subfigure}
    \hfill
    \begin{subfigure}{0.32\textwidth}
        \centering
        \includegraphics[width=\textwidth]{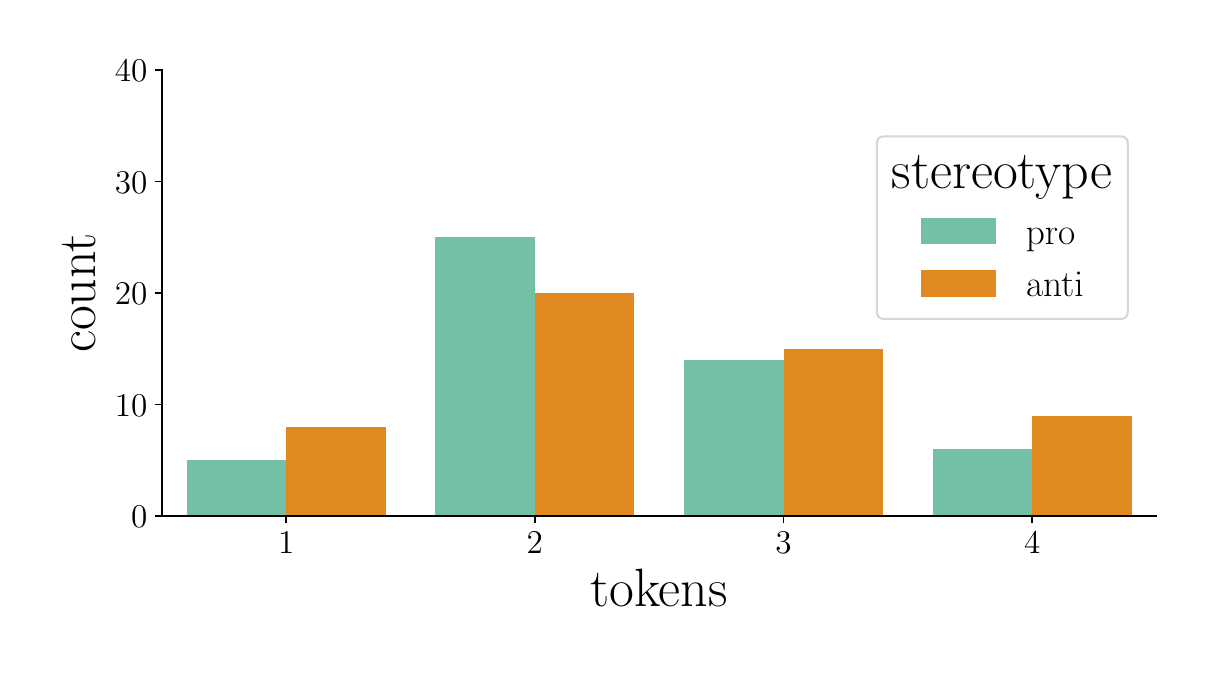}
        \caption{Hebrew}
    \end{subfigure}
    \caption{\textsc{mBART}: Human translated profession names grouped by the number of tokens they were split into. On the x-axis: number of tokens per word. On the y-axis: the count of pro- and anti-stereotypical forms of professions in each of the groups. Pro-stereotypical forms tend to be split into fewer tokens than anti-stereotypical forms.}
    \label{fig:1B_4_mbart}
\end{figure*}

\begin{figure*}[!tb]
    \centering
    \begin{subfigure}{0.32\linewidth}
        \centering
        \includegraphics[width=\textwidth]{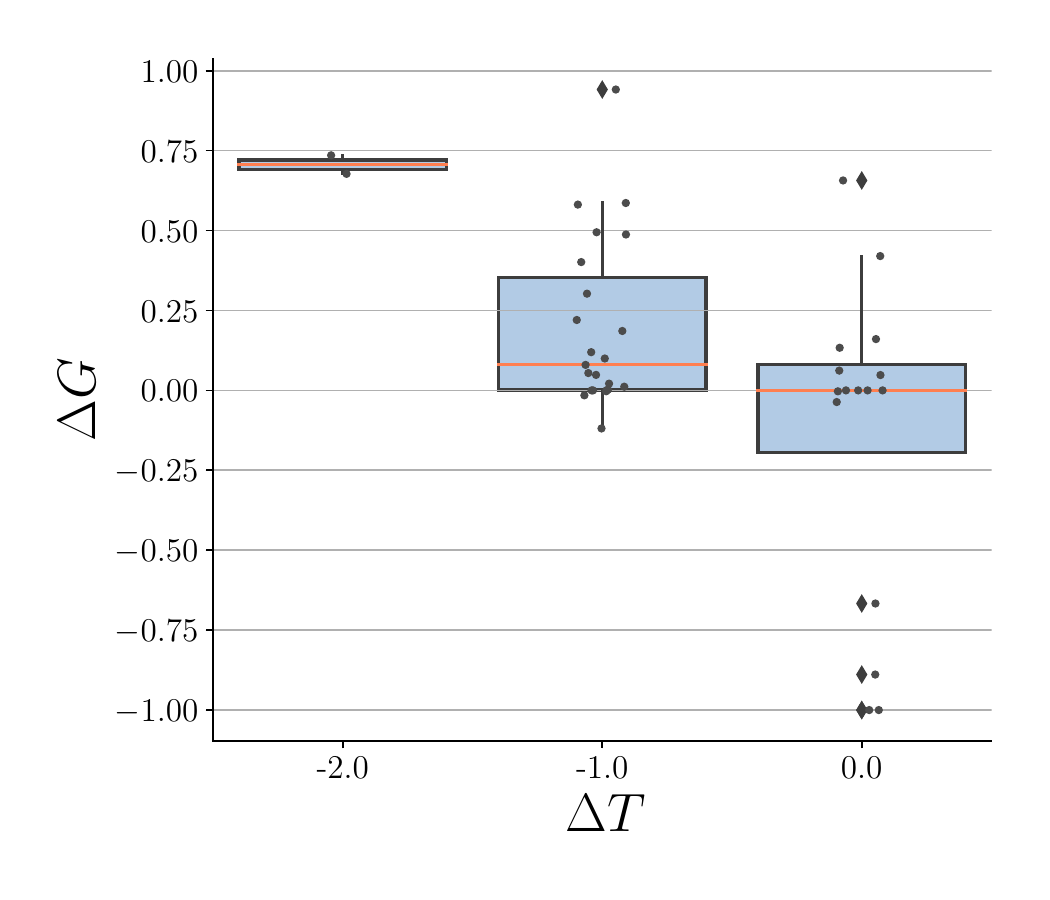}
        \caption{German}
    \end{subfigure}
    \hfill
    \begin{subfigure}{0.32\linewidth}
        \centering
        \includegraphics[width=\textwidth]{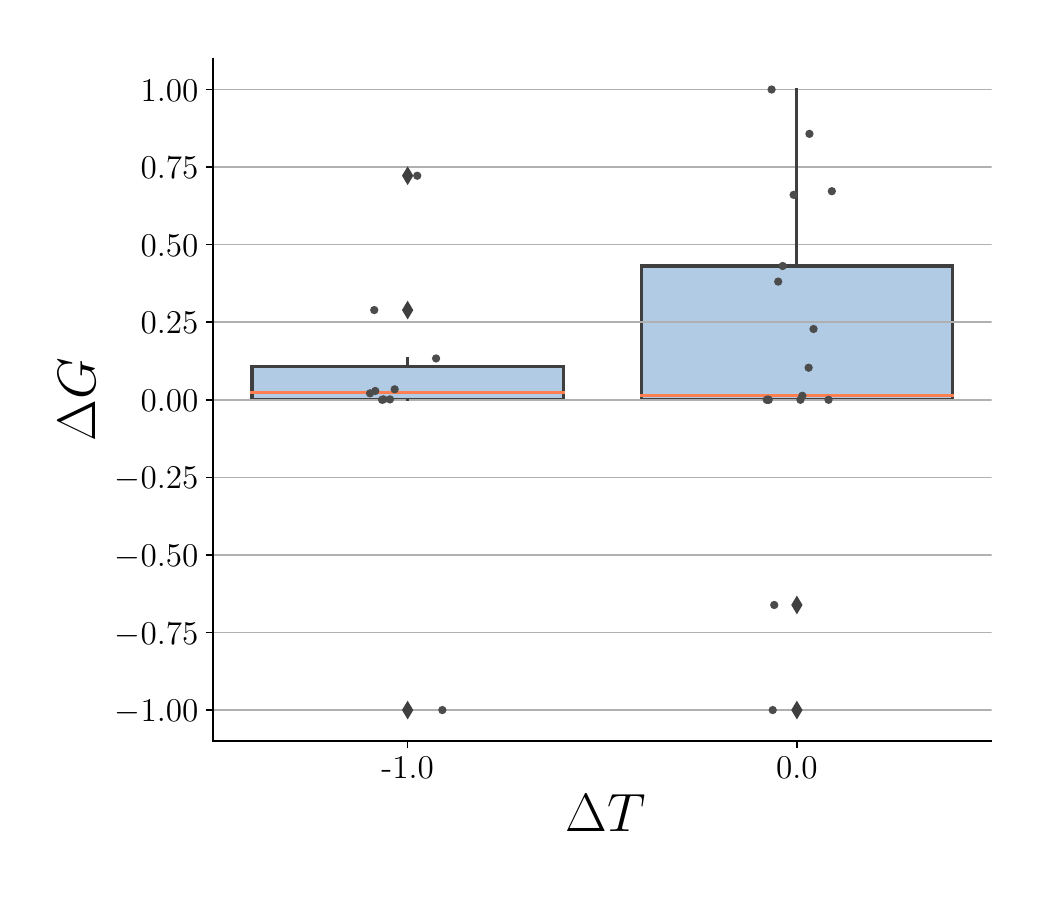}
        \caption{Spanish}
    \end{subfigure}
    \hfill
    \begin{subfigure}{0.32\linewidth}
        \centering
        \includegraphics[width=\textwidth]{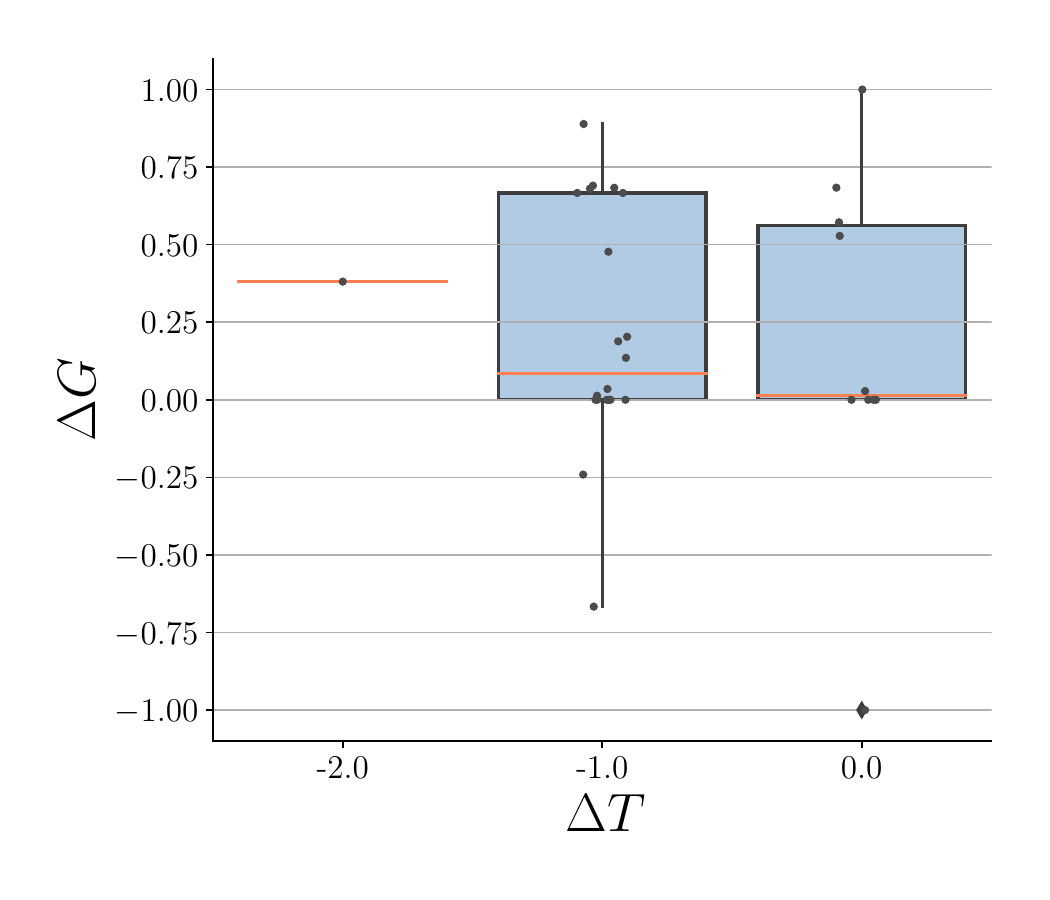}
        \caption{Hebrew}
    \end{subfigure}    
    \caption{\textsc{mBART}: $\Delta G$ as the difference between F1-score for male and female test instances for each of paired translations. $\Delta T$ is the difference between the number of tokens in a male and a female form. The median is marked by an orange line.}
    \label{fig:1C_delta_g_mbart}
\end{figure*}
\begin{figure*}[!tb]
    \centering
    \begin{subfigure}{0.32\linewidth}
        \centering
        \includegraphics[width=\textwidth]{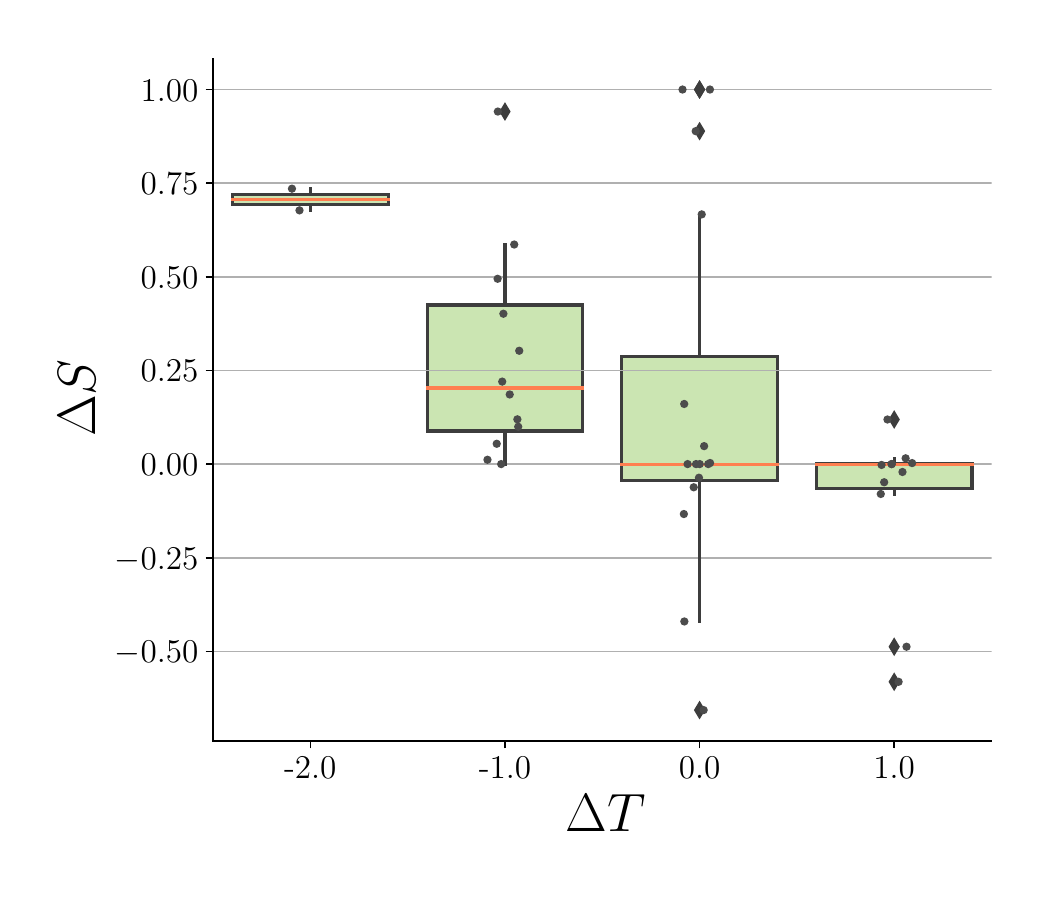}
        \caption{German}
    \end{subfigure}
    \hfill
    \begin{subfigure}{0.32\linewidth}
        \centering
        \includegraphics[width=\textwidth]{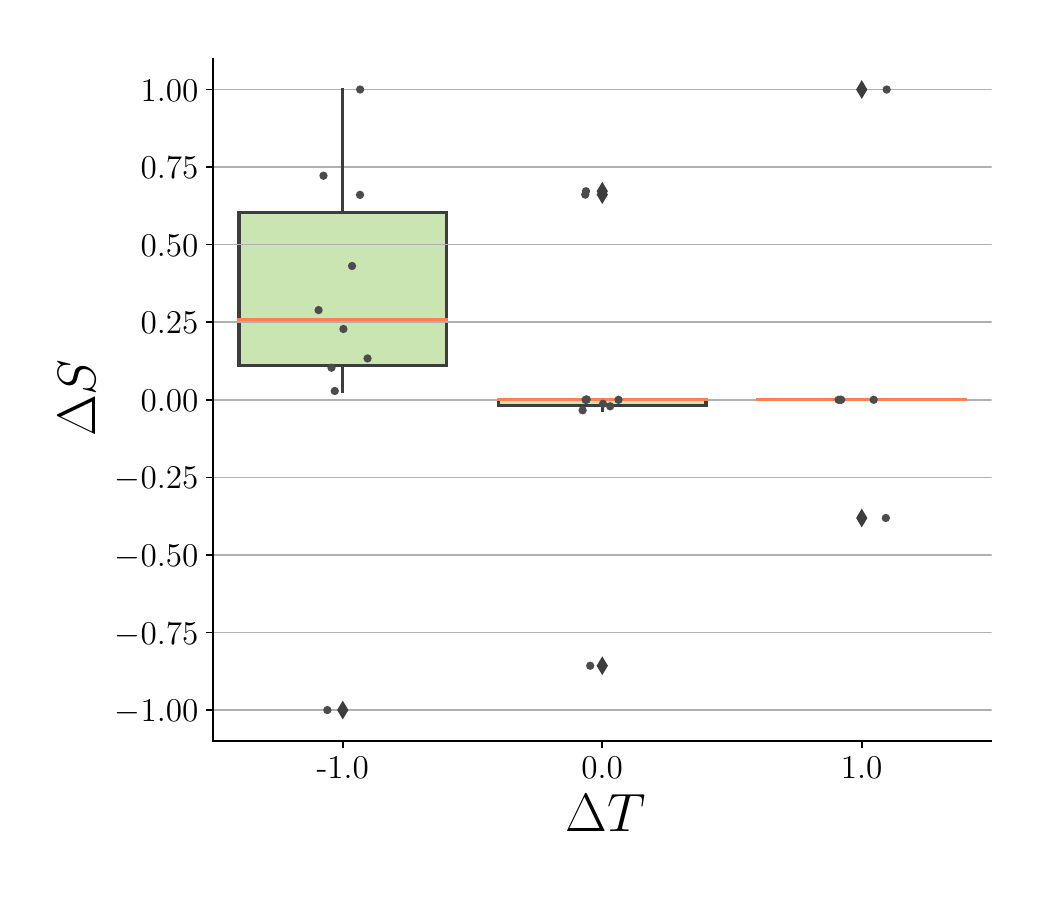}
        \caption{Spanish}
    \end{subfigure}
    \hfill
    \begin{subfigure}{0.32\linewidth}
        \centering
        \includegraphics[width=\textwidth]{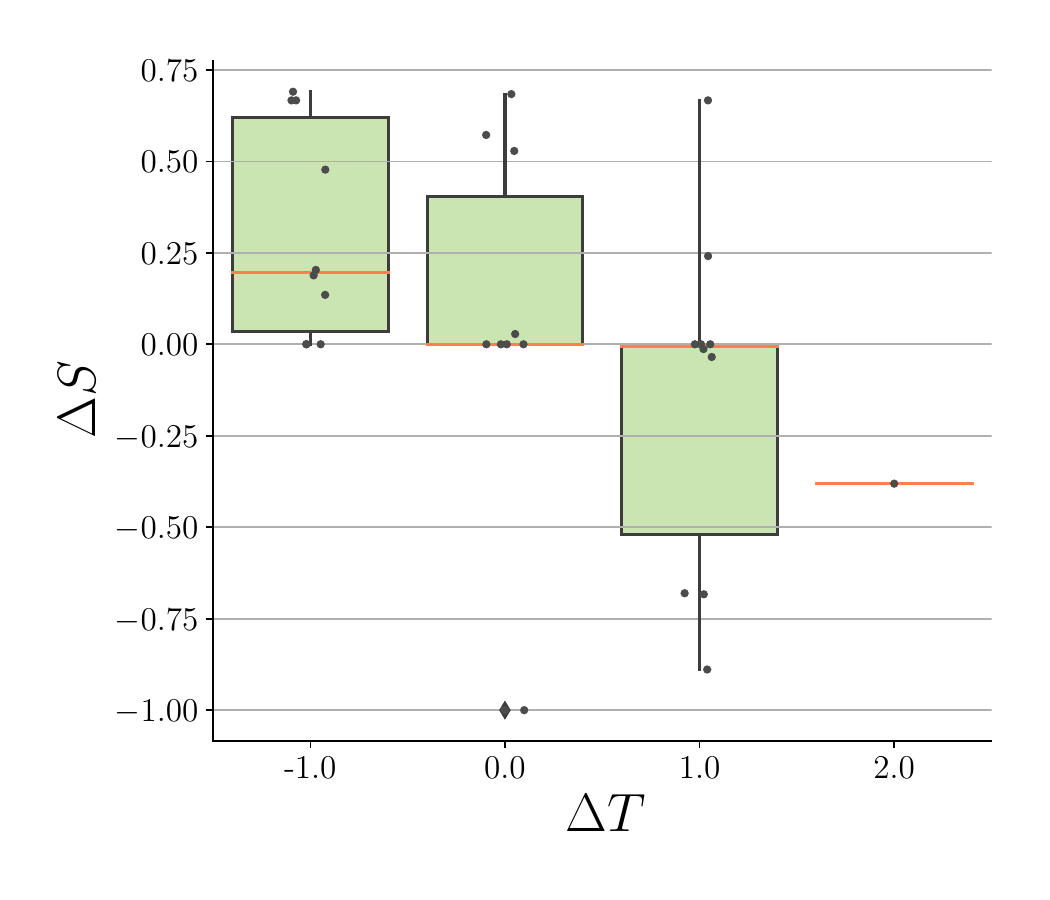}
        \caption{Hebrew}
    \end{subfigure}
    \caption{\textsc{mBART}: $\Delta S$ as the difference between F1-score for pro- and anti-stereotypical test instances for each of paired translations. $\Delta T$ is the difference between number of tokens in a pro- and an anti-stereotypical translation. Median is marked by an orange line.}
    \label{fig:1C_delta_s_mbart}
\end{figure*}

We repeat the analysis of the number of tokens per word described in Section~\ref{sec:1B} for \textsc{mBART}. Figures~\ref{fig:1B_3_mbart}~and~\ref{fig:1B_4_mbart} show that similarly to Opus models, \textsc{mBART} split female and anti-stereotypical words in the target language into more tokens. Noticeably, almost no female words, in all the languages are represented as one token.

Figures~\ref{fig:1C_delta_g_mbart}~and~\ref{fig:1C_delta_s_mbart} show similar trends for $\Delta G$ and $\Delta S$ against $\Delta T$ as observed in Section~\ref{sec:res-1C} for OPUS models.

\section{WinoMT details}
\label{sec:winomt-details}
The WinoMT dataset that we used for evaluation contains 3,888 sentences. Each sentence contains a profession and a pronoun that describes the gender of the profession. The dataset is equally balanced between male and female examples and also between stereotypical and non-stereotypical gender-role assignments. Specifically, there are  1826 sentences with a male pronoun assigned to the profession and 1822 sentences with a female pronoun. The remaining 240 sentences are gender-neutral, i.e., they contain the pronoun ``they''.

Note that this dataset contains only non-ambiguous sentences where the gender of each profession is known. This selection was made to enable us to evaluate the accuracy of the ground-truth gender.


\end{document}